\documentclass[]{article}

\usepackage{amssymb}
\usepackage[T1]{fontenc}
\usepackage{amsmath,amssymb}
\usepackage{xcolor}
\usepackage{todonotes}
\usepackage{url}
\usepackage{changes}
\usepackage{graphicx}
\usepackage{changes}
\usepackage{graphicx}
\usepackage{comment}
\usepackage{array}
\usepackage[labelformat=simple]{subfig}
\usepackage{float}
\usepackage{hyperref}
\usepackage{breakcites}

\usepackage{amsthm}

\theoremstyle{definition}
\newtheorem{assumption}{Assumption}

\newcommand{\set}[1]{\mathcal{#1}}
\newcommand{\Norm}[1]{\mathcal{N}\big(#1\big)}

\newcommand{\expeceta}[1]{\mathbb{E}_\eta \left(#1\right)}

\newcommand{\expec}[1]{\mathbb{E}\left(#1\right)}
\newcommand{\Prob}[1]{\mathbb{P}\left(#1\right)}

\begin{document}

\title{Robust Bayesian Target Value Optimization}
\author{Johannes G. Hoffer\thanks{voestalpine B\"ohler Aerospace GmbH \& Co KG, Mariazellerstra\ss{}e 25, Kapfenberg, Austria}, Sascha Ranftl\thanks{Institute of Theoretical Physics-Computational Physics, Graz University of Technology, 8010 Graz, Austria}, Bernhard C. Geiger\thanks{Know-Center GmbH, Inffeldgasse 13, Graz, Austria}}

\maketitle

\begin{abstract}
  We consider the problem of finding an input to a stochastic black box function such that the scalar output of the black box function is as close as possible to a target value in the sense of the expected squared error. While the optimization of stochastic black boxes is classic in (robust) Bayesian optimization, the current approaches based on Gaussian processes predominantly focus either on i) maximization/minimization rather than target value optimization or ii) on the expectation, but not the variance of the output, ignoring output variations due to stochasticity in uncontrollable environmental variables. In this work, we fill this gap and derive acquisition functions for common criteria such as the expected improvement, the probability of improvement, and the lower confidence bound, assuming that aleatoric effects are Gaussian with known variance. Our experiments illustrate that this setting is compatible with certain extensions of Gaussian processes, and show that the thus derived acquisition functions can outperform classical Bayesian optimization even if the latter assumptions are violated. An industrial use case in billet forging is presented.
\end{abstract}

\section{Introduction}

Inverse problems, where one aims to find parameters of a system either explaining or guaranteeing certain behavior, are ubiquitous in science and industry. Consider, for example, process control in manufacturing, where a part or material undergoes a specific manufacturing process characterized by tuneable design variables. These design variables should be optimized such that the output of the manufacturing process is as close as possible to a defined target. However, manufacturing processes are often influenced by uncertainties of different kinds, such as material imperfections, variation within process tolerances, seasonal effects, or limited accuracy for controlling process variables. These uncertainties, which are often summarized as environmental effects, need to be taken into account when solving inverse problems.

Standard Gaussian processes (GPs) are capable of solving inverse problems under uncertainties, and may comprise distinct kinds of uncertainties~\cite{Ranftl2021,Ranftl2020}, e.g., aleatoric and epistemic uncertainties.
Indeed, several acquisition functions have been proposed for ``noisy''~\cite{Gramacy2010OptimizationUU,Huang2006,Letham2019,Picheny2010}, and ``robust''~\cite{Kirschner2022, Bogunovic2018} Bayesian optimization (BO), cf.~Section~\ref{sec:related}. However, the majority of the previous works on GP-based BO does not distinguish between epistemic and aleatoric uncertainties, which arise from finiteness of training data and stochasticity in the relationship between input and output, respectively. Those works that do either focus on maximization/minimization settings rather than on target value optimization, or their understanding of robustness against aleatoric effects is limited to optimizing the \emph{expected} output of the black box function, ignoring its variance due to aleatoric effects. Indeed, while the acquisition functions for target value optimization in~\cite{pmlr-v89-uhrenholt19a} are not robust against aleatoric effects, the few works that simultaneously try to optimize the expected output and minimize the output variance due to variations in the environmental variables either focus on maximization/minimization~\cite{dai2017stable,Iwazaki2021} or fail to fully exploit the mathematical peculiarities of target value optimization~\cite{hoffer2022gaussian}.

Thus, the literature exhibits a striking and practically relevant gap that this work seeks to fill. Specifically, we will derive acquisition functions for robust Bayesian target value optimization, with the aim of selecting design variables such that the black box function output is close to a target value in the sense of an expected squared error. This aim not only requires that the expected output of the black box function is close to the target, but that also its variation due to aleatoric effects is small. Essentially, our approach is based on a separation between aleatoric (for evaluating the expected squared error) and epistemic (for formulating the acquisition function) uncertainties.

We set out with the assumptions that the aleatoric effects are Gaussian with known variance function and that they can be quantified separately from epistemic effects (Section~\ref{sec:setup}), and we show in our experiments in Section~\ref{sec:experiments} that they approximately hold for certain practically relevant models based on GPs. Based on these assumptions, we derive acquisition functions for target value optimization in Section~\ref{sec:acquisitionfunction}. Specifically, we show that by measuring the quality of the optimization by the squared error expected due to aleatoric effects, that the resulting acquisition functions can be computed in closed form and are reminiscent of those for noise-free target vector optimization~\cite{pmlr-v89-uhrenholt19a}. Using both synthetic and real-world examples, we show in Section~\ref{sec:experiments} in which cases our proposed acquisition functions outperform classical BO even when some of our assumptions are violated. We summarize the insights from these experiments and discuss limitations of our work in Section~\ref{sec:discussion}.

\section{Problem Definition}
\label{sec:setup}
\newcommand{\reals}{\mathbb{R}}
We consider the optimization of a black box function $f{:}\ \reals^{D+A}\to\reals$ that maps a vector $x$ of $D$ controllable design variables and a vector $\eta$ of $A$ uncontrollable environmental variables to a scalar output $y$, i.e., $y=f(x,\eta)$. Our aim is to select the design variables $x$ such that the output $y$ is as close as possible to a target value $y^\bullet$ in the sense of an expected squared error, where the expectation is taken over the unknown environmental variables. Mathematically, we are interested in finding a solution to
\begin{equation}\label{eq:opt}
    \arg\min_x \expeceta{\big(y^\bullet - f(x,\eta)\big)^2} =: \arg\min_x E(x)
\end{equation}
where $\expeceta{\cdot}$ denotes expectation w.r.t.\ $\eta$.
In this setting, the environmental variables $\eta$ correspond to aleatoric effects that cannot be controlled by optimization. Further, the setup is general enough to cover measurement errors ($y=f(x)+\eta$), uncertain inputs to a black box function ($y=f(x+\eta)$), and more complicated settings.

To formulate this optimization problem within the framework of BO and to approach it using GPs, we will introduce three simplifying assumptions.

\begin{assumption}\label{assumption:Gaussian}
 The mapping from the design variables $x$ to the output $y$ is Gaussian, i.e., we assume that it is fully characterized by a mean function $m(x):=\expeceta{f(x,\eta)}$ and a variance function $\sigma_a^2(x):=\expeceta{\big(f(x,\eta)-m(x)\big)^2}$.
\end{assumption}

Assumption~\ref{assumption:Gaussian} allows us to derive analytic expressions for the acquisition functions. Further, under this assumption the optimization objective in~\eqref{eq:opt} simplifies to
\begin{equation}\label{eq:opt:Gauss}
    E(x) = \big(y^\bullet-m(x)\big)^2 + \sigma_a^2(x).
\end{equation}

\begin{assumption}\label{assumption:noise-free}
 For all $x$, we can take noise-free measurements of the mean function $m(x)$, i.e., we can build a dataset $\set{D}=\big\{\big(x_i,m(x_i)\big)\big\}_{i=1,\dots,N}$.
\end{assumption}

Assumption~\ref{assumption:noise-free} essentially requires that we have access to measurements of the black box function for which aleatoric effects have been averaged out. This can be achieved by  i) taking sufficiently many repeated measurements or simulations with the same design variable, approximating the expectation $\expeceta{f(x,\eta)}$ by an average, or by ii) measuring (in addition to $x$ and $y$) and subsequently marginalizing the (uncontrollable, but measurable) environmental effects $\eta$.
This latter approach is common in the literature on robust BO, where GP models are created for $f(x,\eta)$ and marginalized over $\eta$~\cite{Oliveira2020,Iwazaki2021,ToscanoPalmerin2022}.
In particular, \cite{Ankerman2010} found that this is a good approximation for separating 'intrinsic' and 'extrinsic' variances even with very low numbers of repetitions.
Also, e.g.,~\cite{Girard2004} assumes, based on GPs, either noise-free learning or noise-free inference, albeit mainly due to the mathematical complexity of considering noise in both simultaneously. As a consequence of Assumption~\ref{assumption:noise-free}, we can use a GP-based estimate $\hat m$ of $m{:}\ \reals^D\to\reals$ which coincides with $m$ exactly at all values of the design variable $x_i$  that have been measured so far.

A real-world example for such kind of data could be manufacturing chains, where adjusting the design parameters can be very costly while several repetitions for a fixed design parameter set can be cheap. Another real-world example are data from computer simulations, where the design parameters may be related to geometry descriptions, which require elaborate mesh (re-)constructions, or scale parameters in stochastic simulations, which require (re-)tuning of algorithmic parameters such as in Markov Chain Monte Carlo methods.

\begin{assumption}\label{assumption:environmental}
 We have full knowledge of $\sigma_a^2{:}\ \reals^D\to\reals_0^+$. 
\end{assumption}

This last requirement of a fully known aleatoric variance function seems restrictive and appears to limit the practical utility of our approach. However, while we will derive our acquisition functions based on this assumption, our experiments in Section~\ref{sec:experiments} will show that even with crude (learned or computed) estimates $\hat{\sigma}_a^2$ of $\sigma_a^2$ our approach can outperform current alternatives to Bayesian target value optimization.
In the real-world examples mentioned before, such estimates can usually be obtained with reasonable effort.

\section{Related Work}
\label{sec:related}

A plethora of acquisition functions based has been suggested, see e.g. \cite{Shahriari2015,Frazier2018,Brochu2010,Loredo2004,Preuss2021,Ramachandran2019}
\cite{zhan2020expected,zhou2020neural} for an overview.
Also  the need to account for aleatoric effects in BO has been long recognized and is known under ``noisy'' or ``robust BO'', predominantly for maximization/minimization settings (rather than for target value optimization).
For improvement-based acquisition functions such as expected improvement (EI) or probability of improvement (PoI), the inherent stochasticity of the mapping makes it difficult to judge which input $x_t$ in the dataset $\set{D}$ is the previous ``best''. To account for this in a minimization setting, the authors of~\cite{Gramacy2010OptimizationUU} set the previous best input to $\arg\min_x \mu(x)$, where $\mu(x)$ is obtained by training a GP on $\set{D}$. Computing and optimizing the resulting EI in a similar way as in the noiseless case was shown to lead to slow convergence to the minimum of the mapping, cf.~\cite[Sec.~2.1]{Letham2019}. \cite{Huang2006} instead select $x_t$ as the ``effective best solution'' and augment EI by a multiplicative factor that favors inputs with large epistemic uncertainty. \cite{Letham2019} also consider noisy observations (and noisy constraints) when training GPs for BO, but compute and optimize EI on the predictive distribution for the noiseless mapping $f(x)$, rather than for the noisy mapping $y(x)=f(x)+\eta$. The authors of~\cite{Picheny2010} replace the expected value $\mu(x)$ by the quantile of the GP's predictive distribution, and compute EI based on the improvement of this quantile. None of these works, however, accounts for variations of the output due to aleatoric effects.
Further, most of the mentioned studies focus on noise at the output of a deterministic mapping, i.e., $y(x)=f(x)+\eta$. Considering deterministic mappings with noisy inputs, i.e., $y(x)=f(x+\eta)$,~\cite{Girard2004,Girard2003} derive approximations for a GP's predictive posterior mean and variance. These approximations were used in~\cite{dai2017stable} to separate epistemic effects and aleatoric effects propagated through the mapping to derive an acquisition function for the upper confidence bound (UCB) to stabilize BO in the sense of minimizing output variations due to noisy inputs.

Robust BO more generally considers optimization in uncertain environments, where in addition to controllable design variables $x$, the black box function depends also on uncontrollable environmental variables $\eta$, i.e., $y=f(x,\eta)$. In this line of research, \cite{ToscanoPalmerin2022} optimize the expectation of $y$ over $\eta$, i.e., it maximizes the mean function $m(x)$. \cite{Iwazaki2021} extend beyond~\cite{ToscanoPalmerin2022} by proposing confidence bound-based approaches to simultaneously maximize the expectation $m(x)$ and minimize the variance $\sigma_a^2(x)$, cf.~\cite[Fig. 1 and Sec. 3.2]{Iwazaki2021}, while~\cite{Kirschner2022} remain optimizing $m(x)$ but relax the assumption that the distribution of $\eta$ is known. \cite{Daulton_MVAR} maximize the (multivariate) value-at-risk, which implicitly balances maximizing the expectation and minimizing the variance. \cite{Frohlich2020} aims to maximize the expectation $m(x)$ of $y=f(x+\eta)$  using a variant of entropy search. \cite{Nogueira2016} proposed to use the expectation of the unscented expected improvement (UEI) acquisition function with respect to the input noise. \cite{Oliveira2020} aim to maximize a black box function with noise at the input \emph{and} the output. While they also consider noisy measurements during surrogate modeling and optimization,~\cite{Frohlich2020,Nogueira2016,Beland2017,Bogunovic2018} assume exact knowledge of the function input during optimization and only aim to achieve robustness during deployment. Thus, these works make assumptions similar to our Assumption~\ref{assumption:noise-free}, but focus on maximization/minimization and do not study the peculiarities of target value optimization in the light of (aleatoric) input uncertainty.

For target vector optimization, one aims at minimizing a distance $d(x)=d(y(x),y^\bullet)$, e.g., the squared Euclidean distance $\Vert y-y^\bullet \Vert^2_2$, between the mapping's output $y$ and a target value or target vector $y^\bullet$.\footnote{An entirely different approach was proposed in~\cite{Beland2017}, where a GP prior is imposed directly onto an optimality criterion or loss function, in contrast to the more widely-spread procedure of imposing the GP prior on the function which is to be optimized.} In the Gaussian case and for a $k$-dimensional output vector $y(x)$, a non-central $\chi^2$-distribution was shown to be an unbiased estimate for the posterior $p(d(x)|x,\set{D})$ in~\cite{pmlr-v89-uhrenholt19a}. 
For this setting, the authors derived acquisition functions for EI and lower confidence bound (LCB) for both standard GP and warped GP regression, cf.~\cite{snelson2004warped}.   Target value optimization for $y^\bullet=0$ was considered in~\cite{Osborne2009} for a stochastic mapping and additional stochasticity in the input, but this work does not distinguish between epistemic and aleatoric effects. The authors of~\cite{Pandita2016} acknowledge the distinct nature of aleatoric and epistemic uncertainty and quantify the epistemic uncertainty of the optimal location, but they do not quantify or exploit aleatoric uncertainty in their procedure. 
Also \cite{JEONG2021} considered vector-valued outputs and targets and arrived at a similar result as \cite{pmlr-v89-uhrenholt19a}, albeit approaching the problem from a multi-objective perspective. They allowed for different weightings of the different objectives in contrast to \cite{pmlr-v89-uhrenholt19a}, and, in contrast to this work, did not consider robustness with respect to aleatoric effects in stochastic environmental variables.
\cite{Astudill2022,pmlr-v97-astudillo19a} leveraged prior knowledge on composite structures of the objective function similarly as it appears in \cite{pmlr-v89-uhrenholt19a}, albeit again without considering aleatoric input noise.
To the best of our knowledge, the only approach that performs target vector optimization and that considers the variance of the black box function output due to aleatoric effects
is~\cite{hoffer2022gaussian}, where the authors utilize the UCB acquisition function from~\cite{dai2017stable} and where the aleatoric uncertainty is given by the mean function of a separate GP. Their approach, however, ignores the fact that the squaring operation ensures that the posterior of $d(x)$ given $x$ and the dataset $\set{D}$ has non-negative support.

\section{Robust Bayesian Target Value Optimization Using Gaussian Processes}
\label{sec:acquisitionfunction}

Suppose that at optimization iteration $t$ we have access to a dataset $\mathcal{D}=\big\{\big(x_i,m(x_i)\big)\big\}$, $i=1,\dots,t$ of noise-free measurements of the mean function $m$, and that we approximate this mean function by a GP. 
We denote the predictive posterior mean function of this GP as $\mu$ to distinguish it from the real mean function $m$; the predictive posterior variance function of this GP, which estimates the epistemic uncertainty resulting from the finite dataset $\mathcal{D}$, is denoted as $\sigma_{e}^2{:}\ \reals^D\to \reals_0^+$. Thus, our estimate $\hat m$ of $m$ is given by
\begin{equation}\label{eq:GPsurrogate}
\hat m(x) \sim \Norm{\mu(x),\sigma_e^2(x)}
\end{equation}
with
\begin{subequations}\label{eq:GPStandard}
\begin{align}
    \mu(x) &= \mathbf{k}(x)\left(K+\sigma^2 I\right)^{-1}\mathbf{m}^T\\
    \sigma_{e}^2(x) &= k(x,x)-\mathbf{k}(x)\left(K+\sigma^2 I\right)^{-1}\mathbf{k}^T(x)
\end{align}
\end{subequations}
where $\sigma^2$ is measurement noise, $k{:}\ \mathbb{R}^2\to\mathbb{R}$ denotes a kernel function, $\textbf{x}=(x_1,\dots,x_t)$ and $\textbf{m}=\big(m(x_1),\dots,m(x_t)\big)$ are the vectorized dataset $\mathcal{D}$, $[K]_{ij}=k(x_i,x_j)$ is the kernel matrix, and where $\mathbf{k}(x)=\big(k(x_1,x),\dots,k(x_t,x)\big)$ denotes the vector of covariances between $\textbf{x}$ in $\set{D}$ and the candidate input $x$.

Utilizing Assumption~\ref{assumption:noise-free}, i.e., the fact of being noise-free in our design of the GP, we can set the hyperparameter $\sigma^2$ to zero.
As a consequence, for all $x_i\in\set{D}$, we have that the epistemic uncertainty $\sigma_e^2(x_i)=0$~\cite[eq.~(2.19)]{RasmussenWilliams}, and that $\hat m(x_i)\equiv\mu(x_i)=m(x_i)$. 
For general $x$, our surrogate $\hat y(x)$ of the black box function output $y(x)$ is Gaussian with mean $\mu(x)$ and variance $\sigma_a^2(x)+\sigma_{e}^2(x)$.
Summarizing, we have
\begin{subequations}
\begin{align*}
    y(x)=f(x,\eta)\sim\Norm{m(x),\sigma_a^2(x)} & \quad \text{Assump.~\ref{assumption:Gaussian}}\\
    \hat m(x) \sim \Norm{\mu(x),\sigma_e^2(x)} & \quad \text{eq.~\eqref{eq:GPsurrogate}}\\
    \forall x_i\in\set{D}{:}\ \sigma_e^2(x_i)=0 & \quad \text{Assump.~\ref{assumption:noise-free}}\\
    \forall x_i\in\set{D}{:}\ \mu(x_i)=m(x_i) & \quad \text{Assump.~\ref{assumption:noise-free}}\\
    \hat y(x) \sim \Norm{\hat m(x),\sigma_a^2(x)} & \quad \text{Assump.~\ref{assumption:environmental}}\\[3pt]
    \hline \hline \\[-10pt]
    \Rightarrow 
    \hat y(x) \sim\Norm{\mu(x),\sigma_a^2(x)+\sigma_e^2(x)}
\end{align*}
\end{subequations}
The first line in this summary should denote that the output $y$ of the black box function $f$ is Gaussian for given $x$, subsuming the stochasticity of environmental variables $\eta$ in $f$.

We now propose acquisition functions to optimize this surrogate. Specifically, and connecting to~\eqref{eq:opt}, we aim to select $x$ such that $\hat y(x)$ is as close as possible to a target value $y^\bullet$. Specifically, we aim to find an input $x$ such that the expected squared error
\begin{equation}
    \hat E(x) := \big(\hat m(x)- y^\bullet\big)^2 + \sigma_a^2(x)
\end{equation}
is minimized. Note that due to the epistemic uncertainty resulting from the finite dataset $\mathcal{D}$, $\hat E(x)$ is a random variable unless $x\in\set{D}$.
Thus, while $E(x)$ from~\eqref{eq:opt} is deterministic since it is an expected value, $\hat E(x)$ is estimated from the GP surrogate of the black box function and is hence random due to the inherent epistemic uncertainty.
Indeed, for $x\notin \set{D}$ the normalized expected squared error $e(x)$ follows a non-central $\chi^2$-distribution:
\begin{subequations}
\begin{equation}
   e(x):= \frac{\hat E(x)-\sigma_a^2(x)}{\sigma_{e}^2(x)} \sim NC\chi^2\big(K=1,\lambda(x)\big),
\end{equation}
with $K=1$ degree of freedom and non-centrality parameter
\begin{equation}\label{eq:lambda}
    \lambda(x)=\frac{\big(\mu(x)-y^\bullet\big)^{2}}{\sigma_{e}^2(x)}.
\end{equation}
\end{subequations}
For $x_i\in\set{D}$, we have $\mu(x_i)=m(x_i)$ and $\sigma_e^2(x_i)=0$, hence $\hat E(x_i)=E(x_i)$. On the one hand, this justifies optimizing 
with
the surrogate instead of the black box function,
On the other hand, it follows that for these inputs, $\hat E(x_i)$ is deterministic. Further, the normalized expected squared error either evaluates to $e(x_i)=0$ if $m(x_i)=y^\bullet$, or to $e(x_i)=\infty$ otherwise.
 
To prevent numerical issues, in our experiments we set the hyperparameter $\sigma^2$ in~\eqref{eq:GPStandard} to a small, positive number, yielding $\sigma_e^2(x)>0$ for all $x$. If $\sigma^2\ll\sigma_a^2(x)$, one still has $\hat E(x_i)\approx E(x_i)$ for all $x_i\in\set{D}$ and, thus, a valid surrogate model for optimization. In the remainder of this section, however, we will stick to $\sigma^2=0$ and to Assumption~\ref{assumption:noise-free}.

\subsection{Improvement-Based Acquisition Functions}
Let without loss of generality $x_t=\min_{x_i\in\mathcal{D}} \hat E(x_i)$ be the input that minimizes the expected squared error over all previous inputs. By Assumption~\ref{assumption:noise-free}, 
$E_{\min}:=\hat E(x_t)$ is deterministic and evaluates to
\begin{equation}
\label{eq:E_min}
    E_{\min}=\hat E(x_t) = E(x_t) = \big(m(x_t)-y^\bullet\big)^2 + \sigma_a^2(x_t).
\end{equation}

\paragraph{Probability of Improvement.}
We can select $x_{t+1}$ such that the PoI over $E_{\min}$ is maximized, i.e., we solve
\begin{equation}
    x_{t+1}=\arg\max_x \Prob{\hat E(x)\le E_{\min}-\zeta}
\end{equation}
where $\zeta\ge0$ is a parameter that prefers larger, improbable improvements over more probable, but small improvements. Given that $e(x)$ has a non-central $\chi^2$-distribution with cumulative distribution function (CDF) $F_{1,\lambda(x)}$, the PoI is given by
\begin{align}
    \Prob{\hat E(x)\le E_{\min}-\zeta} =
     \Prob{e(x)\le \frac{E_{\min}-\zeta-\sigma_a^2(x)}{\sigma_{e}^2(x)}}\notag\\
    = F_{1,\lambda(x)}\left(\frac{\big(m(x_t)-y^\bullet\big)^2 + \sigma_a^2(x_t)-\zeta-\sigma_a^2(x)}{\sigma_{e}^2(x)}\right) \label{eq:poi}
\end{align}
where $\lambda(x)$ is given in~\eqref{eq:lambda} and $E_{\min}$ was substituted by~\eqref{eq:E_min}.

\paragraph{Expected Improvement.}
We can also select $x_{t+1}$ such that the EI over $E_{\min}$ is maximized, i.e., we solve
\begin{equation}
    x_{t+1}=\arg\max_x \expec{\max\big(0,E_{\min}-\hat E(x)\big)}
\end{equation}
where $\expec{\cdot}$ takes the expectation w.r.t.\ epistemic uncertainty. By the linearity of expectation and with $e_{\min}(x)=\big(E_{\min}-\sigma_a^2(x)\big)/\sigma_{e}^2(x)$ this can be rewritten as
\begin{align}\label{eq:EI_alea}
    &\expec{\max\big(0,E_{\min}-\hat E(x)\big)} \notag\\
    &= \sigma_{e}^2(x) \expec{\max{\big(0,e_{\min}(x)-e(x)\big)}} \notag\\
    &= \sigma_{e}^2(x) \int_0^{e_{\min}(x)} \left(e_{\min}(x)-e\right) f_{1,\lambda(x)}(e) \mathrm{d}e
\end{align}
where $f_{1,\lambda(x)}$ is the probability density function of the non-central $\chi^2$-distribution. Similarly as in ~\cite[Sec.~3.2]{pmlr-v89-uhrenholt19a}, this integral can be computed in closed form as
\begin{multline}\label{eq:EI}
    \frac{\expec{\max{\big(0,E_{\min}-\hat E(x)\big)}}}{\sigma_{e}^2(x)} = e_{\min}(x) F_{1,\lambda(x)}\big(e_{\min}(x)\big)\\ - F_{3,\lambda(x)}\big(e_{\min}(x)\big) + \lambda(x)F_{5,\lambda(x)}\big(e_{\min}(x)\big)
\end{multline}

\subsection{Lower Confidence Bound}
We can also select $x_{t+1}$ such that the $q$-quantile of $\hat E(x)$ is minimized. As before, we have a non-central $\chi^2$-distribution for $e(x)$.
Thus, the $q$-quantile of the l.h.s.\ of above equation is given by the inverse CDF, $F^{-1}_{1,\lambda(x)}(q)$. As a consequence, we obtain
\begin{equation}\label{eq:LCB}
    x_{t+1} = \arg\min_x \Big( \sigma_{e}^2(x)F^{-1}_{1,\lambda(x)}(q) + \sigma_a^2(x) \Big)\;,
\end{equation}
where the quantile $q$ in $F^{-1}_{1,\lambda(x)}(q) =: \beta$ might be considered a  tuning parameter.

\section{Experiments}
\label{sec:experiments}
We evaluate our acquisition functions for robust Bayesian target value optimization in three experimental settings, where in each we make specific and realistic assumptions regarding the aleatoric uncertainty due to environmental variables $\eta$. In experiment 1, $\eta$ represents aleatoric uncertainty at the output of the black box function, i.e., $y(x)=f(x)+\eta$. In experiment 2, $\eta$ represents aleatoric effects on the inputs, i.e., $y(x)=f(x+\eta)$; further, we relax Assumption~\ref{assumption:environmental} by replacing $\sigma_a^2$ with an estimate $\hat\sigma_a^2$ computed from the respective GP surrogate. Finally, in experiment 3 both the GP surrogate and the estimate $\hat\sigma_a^2$ are learned from noisy data, i.e., Assumptions~\ref{assumption:noise-free} and~\ref{assumption:environmental} are both relaxed. We compare the performance of our acquisition functions to those suggested by classical BO (for target value optimization), where aleatoric uncertainties are not treated separately.

\subsection{Experiment 1}
\label{sec:example1}

\begin{figure*}

\centering
\subfloat[$\sigma_a= 0.01$ \label{fig:ex_1_EI:a}]{\includegraphics[width=0.33\textwidth]{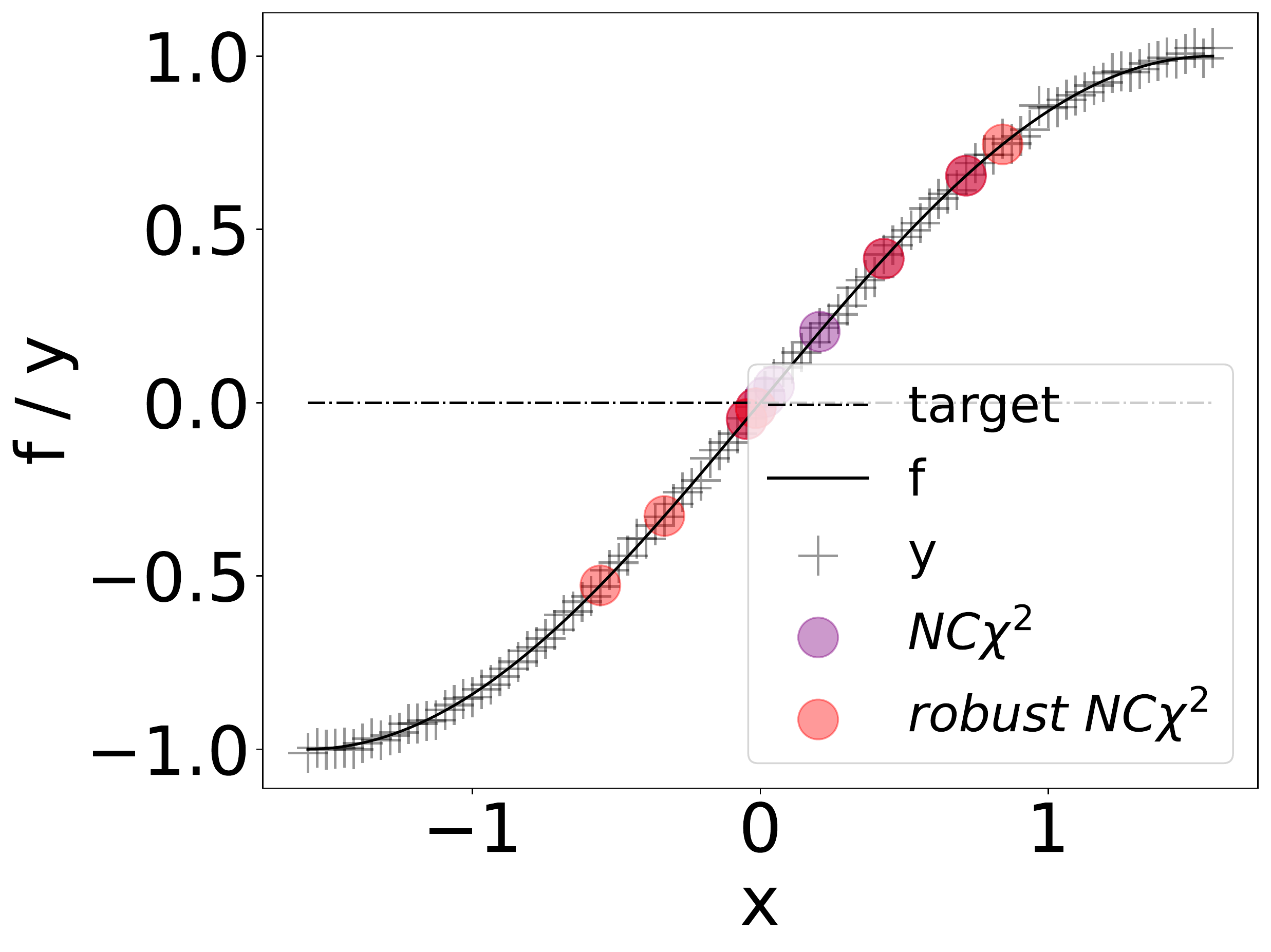}}\hfil
\subfloat[$\sigma_a= 0.1$ \label{fig:ex_1_EI:b}]{\includegraphics[width=0.33\textwidth]{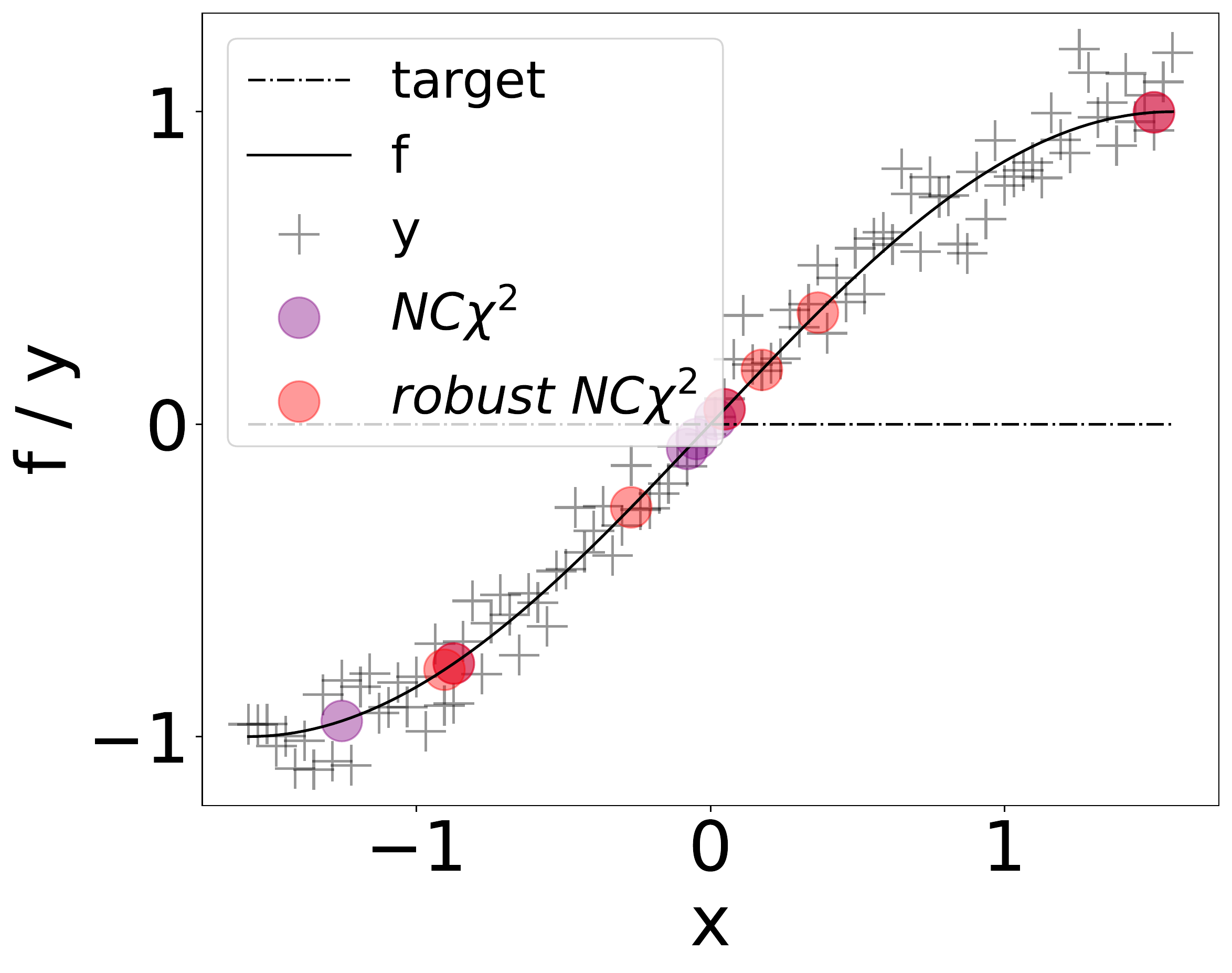}}\hfil 
\subfloat[$\sigma_a= 0.5$ \label{fig:ex_1_EI:c}]{\includegraphics[width=0.33\textwidth]{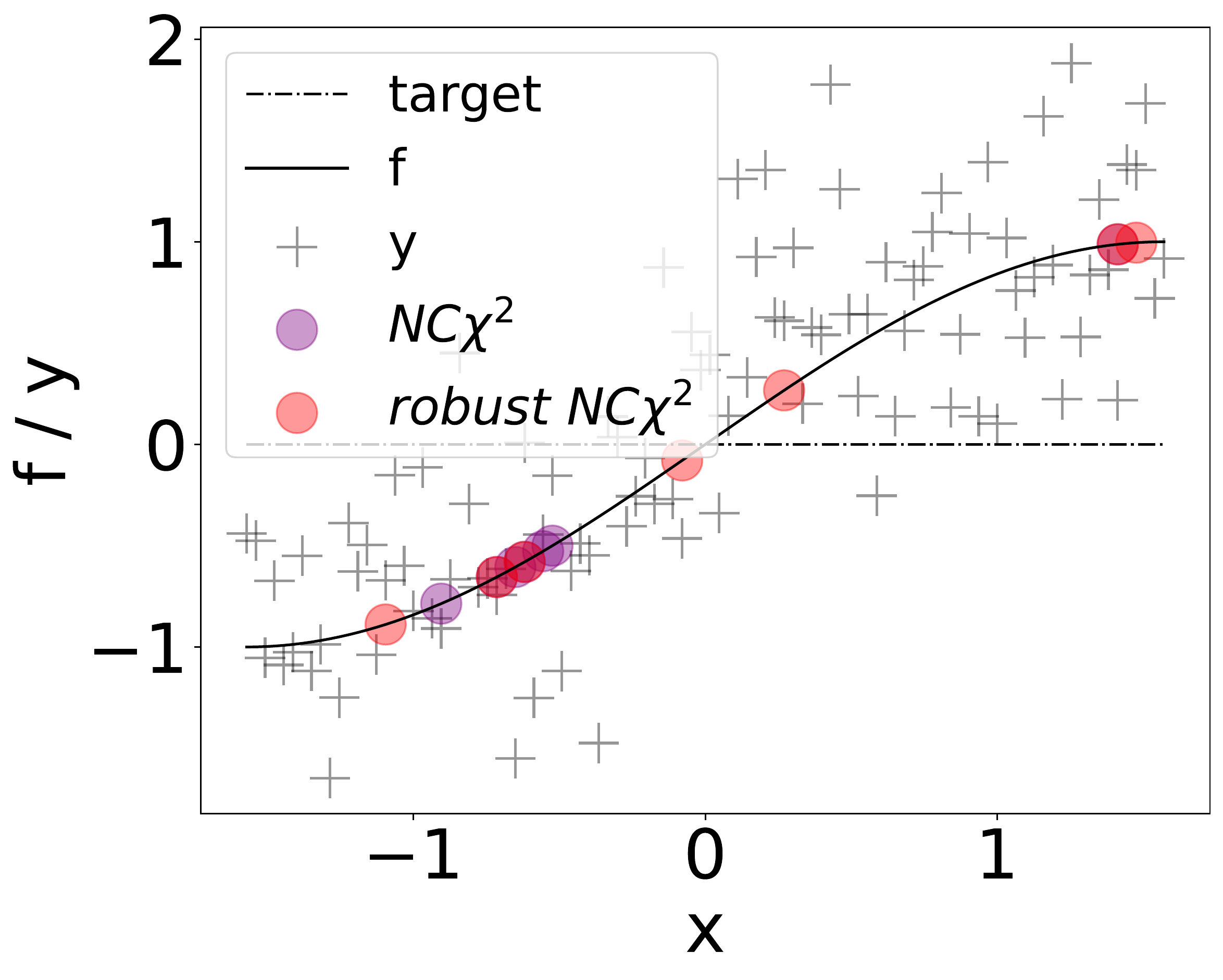}} 

\subfloat[$\sigma_a= 0.01$ \label{fig:ex_1_EI:d}]{\includegraphics[width=0.33\textwidth]{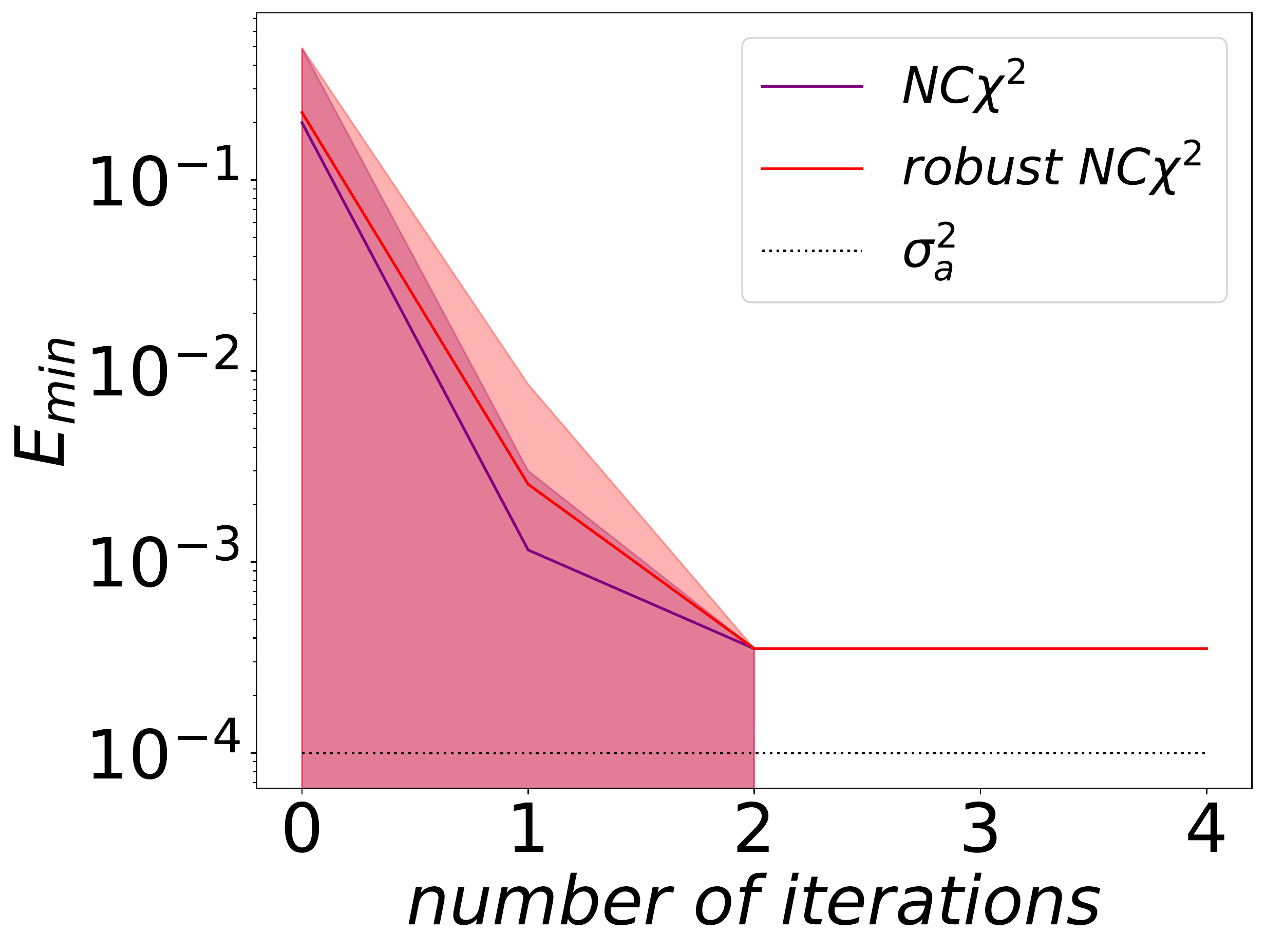}}\hfil   
\subfloat[$\sigma_a= 0.1$ \label{fig:ex_1_EI:e}]{\includegraphics[width=0.33\textwidth]{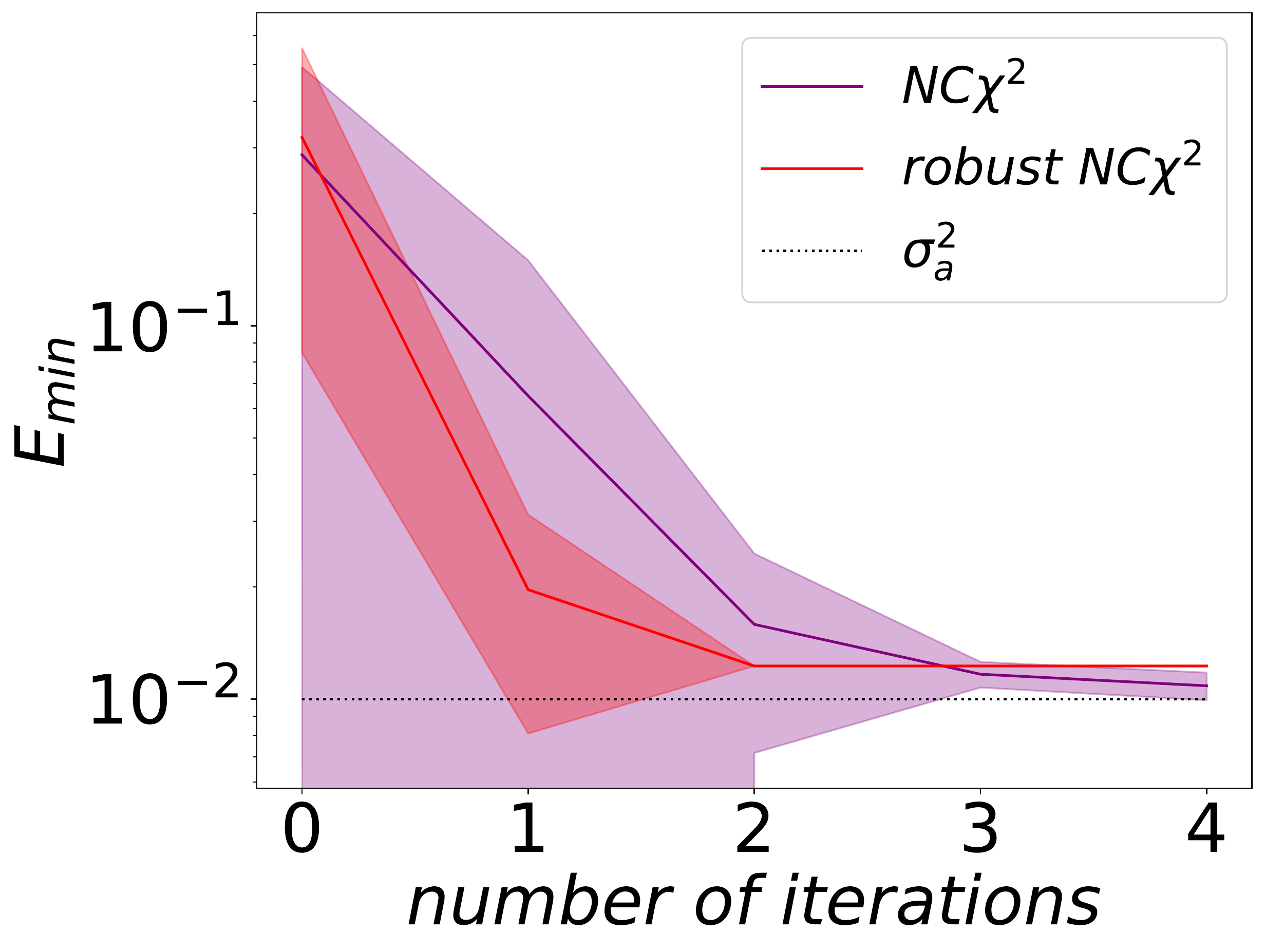}}\hfil
\subfloat[$\sigma_a= 0.5$ \label{fig:ex_1_EI:f}]{\includegraphics[width=0.33\textwidth]{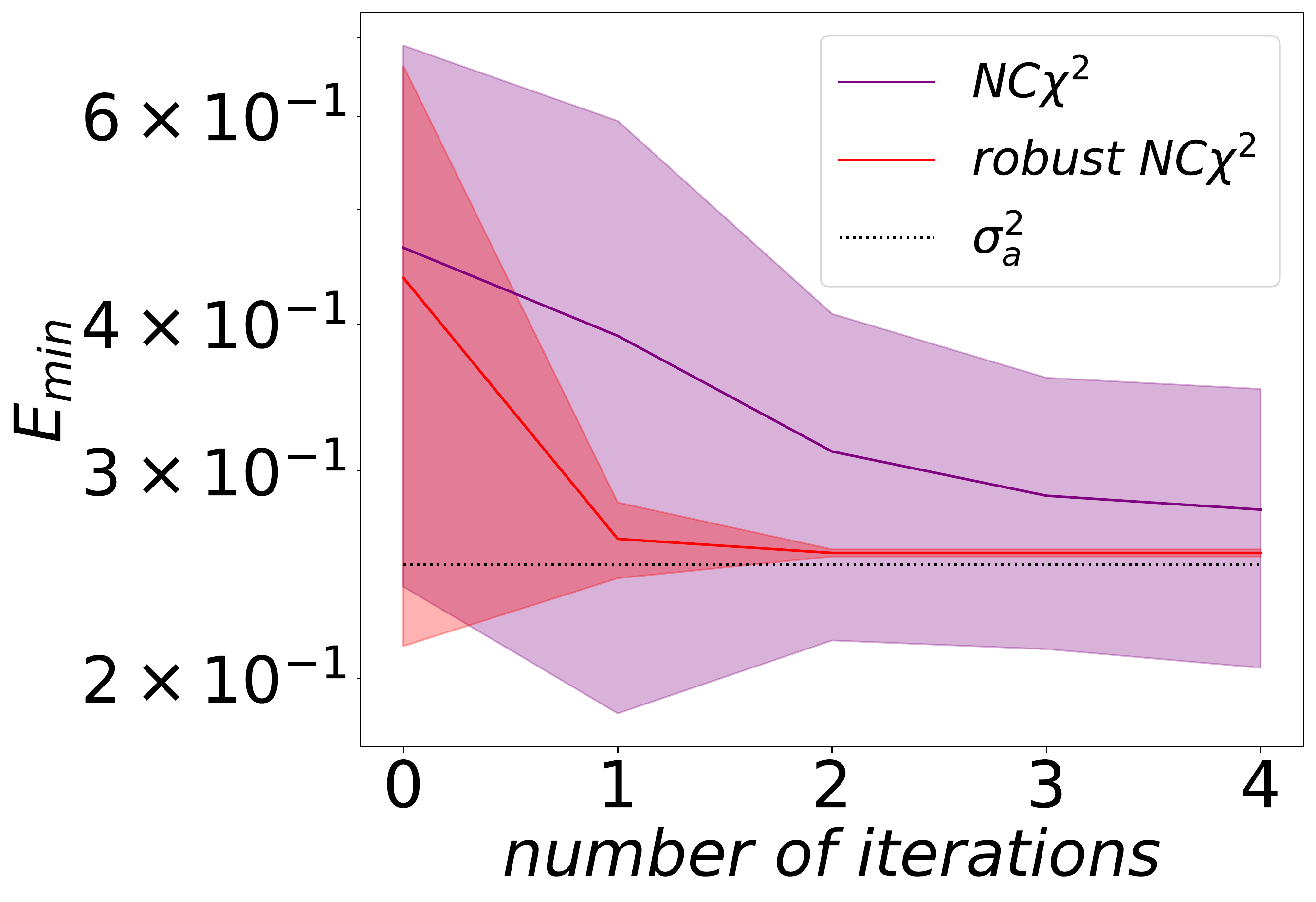}}
\caption{Experiment 1 with EI~\eqref{eq:EI}. (\textbf{a})-(\textbf{c}) Scatterplots for a single run with different levels of aleatoric variance $\sigma_a$. Solid lines represent the noise-free function $f$, crosses the observations $y$, colored dots the sampled candidates for $x$, and dash-dotted lines the target.  (\textbf{d})-(\textbf{f}) $E_{\min}$ (mean and standard deviation over 10 folds) over the number of sampled candidate positions, i.e., number of optimization iterations, for different levels of aleatoric variance $\sigma_a$.}\label{fig:ex_1_EI}
\end{figure*}

In the first experiment, we consider a synthetic stochastic mapping with noise at its output. Specifically, let $y(x)=\sin(x)+\eta$, $x\in[-\pi/2,\pi/2]$ where $\eta\sim\mathcal{N}(0,\sigma_a^2)$ represents the homoskedastic aleatoric uncertainty of the process (see Fig.~\ref{fig:ex_1_EI:a}-\ref{fig:ex_1_EI:c}). To achieve the separation between aleatoric and epistemic uncertainties required in our setting, we train a GP with RBF kernel on noise-free data $\mathcal{D} = \big\{\big(x_i, \sin(x_i)\big)\big\}$. While we have $\sigma_a^2(x)=\sigma_a^2$, we obtain the mean function and the epistemic uncertainty from~\eqref{eq:GPStandard} for $\sigma^2=10^{-10}$, i.e., we use a small, positive number to prevent numerical issues. We compare our approach with the work of \cite{pmlr-v89-uhrenholt19a}, where we train a GP on noisy data $\{(x_i, y_i)\}$ using $\sigma^2 = \sigma_a^2$ in~\eqref{eq:GPStandard}. Both GPs are initially trained on two randomly drawn data points. To account for the randomness of the initial training data points, the mean and standard deviation of all obtained experimental results are calculated and reported in the evaluation, see Figure \ref{fig:ex_1_EI:d} - \ref{fig:ex_1_EI:f}.

Our aim is to find $x$ such that $y$ is as close to $y^\bullet=0$ as possible. 
We utilize our EI acquisition functions (denoted as ``robust $NC\chi^2$'') from Section~\ref{sec:acquisitionfunction} and compare it to the EI acquisition functions from~\cite{pmlr-v89-uhrenholt19a} (denoted as ``$NC\chi^2$''), for different values of $\sigma_a^2$. 
These acquisition functions are evaluated on a fixed, evenly spaced grid of 100 candidate positions in $[-\pi/2,\pi/2]$. Experiments were evaluated for $\sigma_a \in \{0.01,0.1,0.5\}$.

We present the sampled data points of one cross-validation step in Fig.~\ref{fig:ex_1_EI:a}-\ref{fig:ex_1_EI:c} and the minimum expected squared error $E_{\min}$ over the number of optimization steps in Fig.~\ref{fig:ex_1_EI:d}-\ref{fig:ex_1_EI:f}. For low aleatoric uncertainty ($\sigma_a = 0.01$), our acquisition function performs similarly as the one proposed in~\cite{pmlr-v89-uhrenholt19a}, see Fig.~\ref{fig:ex_1_EI:d}. This is expected, as the influence of $\sigma_a$ is rather small in $E_{\min}$, and both approaches coincide for $\sigma_a\to 0$. For increasing $\sigma_a$, a clear advantage w.r.t.\ the convergence of $E_{\min}$ towards $\sigma_a^2$ can be observed, see Figs.~\ref{fig:ex_1_EI:e}-\ref{fig:ex_1_EI:f}. Indeed, our approach samples inputs resulting in outputs close to the target much earlier during optimization, see Figs.~\ref{fig:ex_1_EI:b}-\ref{fig:ex_1_EI:c}. Concretely, given the initial training data points marked in red in Fig.~\ref{fig:ex_1_EI:c}, our acquisition function immediately suggests a position close to the zero crossing (and subsequent draws visible in Fig.~\ref{fig:ex_1_EI:c} simply reduce epistemic uncertainty without affecting $E_{\min}$, cf. Fig.~\ref{fig:ex_1_EI:f}).
This is as expected, since our GP is trained on noise-free data and since our acquisition function makes explicit use of $E(x)$ and $E_{\min}$, respectively.

\subsection{Experiment 2}
\label{sec:example2}
In the second experiment, we consider a synthetic stochastic mapping with no measurement noise, but with noisy inputs. Formally, $y=f(x+\eta)$ with $u = x+\eta$ and  $\eta\sim\mathcal{N}(0,\sigma_u^2)$.
This scenario constitutes a special case of a GP with uncertain inputs, for which, e.g.,~\cite{Girard2003,dai2017stable,McHutchon2011} provided closed expressions for the aleatoric $\sigma_a^2(x)$ and the epistemic $\sigma_{e}^2(x)$ contributions to the predictive variance. 
In other words, \cite{Girard2003} propagated, approximately, the input uncertainty, defined by $\sigma_u^2$, through a GP to the output. In this experiment, the GP surrogate and this approximation is used in order to estimate the aleatoric uncertainty at the output $\hat\sigma_a^2(x) \approx \sigma_a^2(x)$ for a finite $\sigma_u^2$.
Again we set $\sigma^2=10^{-10}$ in \eqref{eq:GPStandard}.

Again we aim at target value optimization, with a target value of $y^\bullet = 0$. We design an illustrative test function $f$, and let $f$ have a comparatively flat region (i.e. small gradient) in the vicinity of the target, where the target itself lies in (or in vicinity of) a region of a steep gradient 

\begin{figure*}[htbp]
\centering
\subfloat[{} \label{fig:ex_2_EI:a}]{\includegraphics[width=0.49\textwidth]{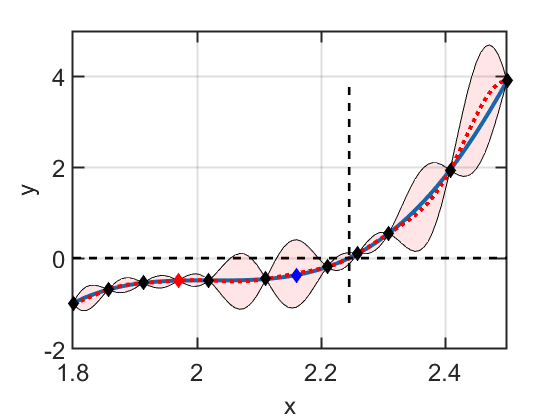}}\hfil

\subfloat[{} \label{fig:ex_2_EI:b}]{\includegraphics[width=0.49\textwidth]{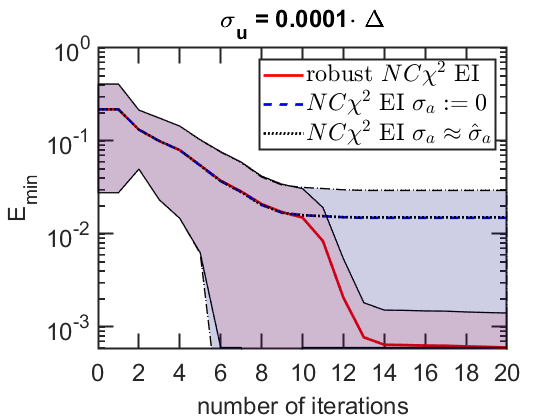}}\hfil 
\subfloat[{} \label{fig:ex_2_EI:c}]{\includegraphics[width=0.49\textwidth]{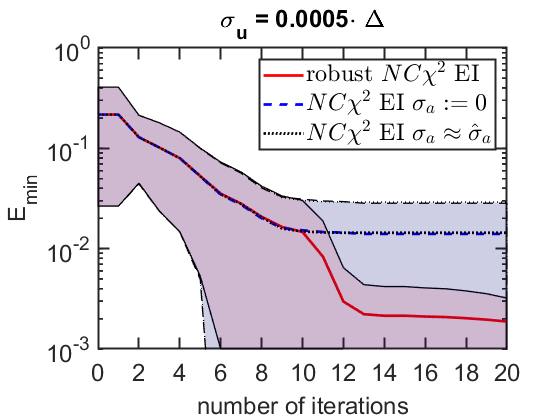}} 
\caption{Experiment 2 with EI. Panel (\textbf{a}) shows the test function $f(x)$ at iteration 10. Light blue: true test function. Black diamonds: data probed in an exemplary optimization loop. Red diamond: Current best value acc. to \eqref{eq:opt}. Blue diamond: Latest proposed next query point. Black dashed: target. Red line and shade: GP mean and uncertainty. Note that the optimum \eqref{eq:opt} differs from the target due to stochastic inputs. Panels  (\textbf{b}) and (\textbf{c}) show the convergence for $f$ (from panel (\textbf{a})) over the number of sampled candidate positions, i.e., number of optimization iterations, for different levels of aleatoric variance in environmental inputs, $\sigma_u$. Red corresponds to our acquisition function \eqref{eq:EI}. Dark blue and black denote the reference methods, \cite{pmlr-v89-uhrenholt19a} and \cite{dai2017stable}, respectively, both detailed in Sec.~\ref{sec:example2}.}\label{fig:ex_2_EI}
\end{figure*}

This is fulfilled by $f(x) := q(x-2)^3 + p/((x-3)^2+r^2) + sx +t$ on $x\in [1.8,2.5]$ with $q =50$, $p=-1$, $r=0.1$, $s=2$, and $t=-3.5$. 
The test function is shown in Fig.~\ref{fig:ex_2_EI:a}.
Note that the scale of ``flat'' and ``steep'' here is determined by the variance of the input, $\sigma_u^2$. Thus we set the parameter as $\sigma_u$ in units of $\Delta$, where $\Delta$ is the interval length of the domain of test function $f$. Results are compared for various values of $\sigma_u$. Note that input uncertainties propagated to the output of a GP can become large quickly, particularly when highly non-linear functions are being modelled like here. This necessitates relatively low values for $\sigma_u$ in order to enable usable GP surrogates in the first place. 
For the training of the GP we use noise-free data $\mathcal{D} =\{x_i, f(x_i)\}$, as assumed previously, inputs for prediction at new inputs remain noisy. Extensions to noisy training inputs $u_i$  may be derived from the findings of~\cite{Girard2004}.

We use an initial data set of two  random data points, and the experiment is averaged over 1000 random repetitions. For the GP, we use an RBF kernel. The acquisition function is evaluated on 100 equidistant candidate positions $x$ in the abovementioned intervals.
I.e. the acquisition function is optimized with a simple grid search.
We compare the performance of: i) our EI acquisition function~\eqref{eq:EI}, which exploits separated aleatoric uncertainty explicitly for target value optimization ($robust~NC\chi^2~EI$),
to ii) the standard (non-robust) EI acquisition function, adapted for target value optimization but ignoring aleatoric uncertainty altogether ($NC\chi^2~EI, \sigma_a:=0$) as proposed by \cite{pmlr-v89-uhrenholt19a}.
We further compare iii) a (non-robust) variant of (ii) where the GP surrogate is corrected for aleatoric uncertainty as in \cite{dai2017stable} ($NC\chi^2~EI, \sigma_a:=\hat\sigma_a$). In other words, variant (iii) is a naive combination of the $NC\chi^2$-distribution for target value optimization from \cite{pmlr-v89-uhrenholt19a} with stability-inducing terms from the Gaussian assumptions in \cite{dai2017stable}.
%

The results for two different noise levels are shown in Fig.~\ref{fig:ex_2_EI:b} and \ref{fig:ex_2_EI:c}. The results demonstrate that our acquisition function~\eqref{eq:EI} (red), leveraging computed estimates of the aleatoric variance, can perform better than equivalent procedures that either do not distinguish aleatoric and epistemic uncertainty (black), or neglect the aleatoric uncertainty (blue) in the first place. In our examples, the state of the art only poorly manages the optimization task. 

This could be attributed to a rather high output uncertainty of the underlying GP due to even comparatively low input uncertainty. This becomes particularly important in the vicinity of highly non-linear behaviour, and considerations for and correction of the aleatoric input uncertainty become relevant.
In Fig.~\ref{fig:ex_2_EI}, both naive approaches behave similarly and get stuck in an apparent local optimum that is close to the target, however in a region where the function derivative is large, leading to a large expected error due to input noise. In contrast, our acquisition function converges towards an optimum that is more stable - at the expense of moving slightly away from the original target value.

\subsection{Experiment 3}
\label{sec:example3}

In our third experiment, we utilize a use case in the field of manufacturing: preforming a super alloy billet on a forging machine.
In forging, outputs must be within a pre-defined tolerance range, w.r.t. control measurements. 
If a forged part is out of tolerance it has to be reworked, e.g., machined, or in extreme cases, scrapped. 
Due to the fact that a process cannot be fully controlled, there is aleatoric variance present, which has to be considered to be within tolerance.
The use case addresses the problem of finding ideal input variables to achieve an output that minimizes $E_{min}$, w.r.t. a chosen output target.

The inputs $x$ for this manufacturing process are process-related, such as setting values $\theta_{1}$ and $\theta_{2}$ for the forging machine, and material-related, such as the initial billet diameter $d$ and height $h$, and the billet temperature $T$. We treat these inputs jointly, i.e., $x = (\theta_{1}, \theta_{2}, d, h, T)$.
The maximum billet diameter $y(x)=d_{\max}(x)$ is the output of the process and is subject to heteroskedastic aleatoric uncertainty with unknown variance $\sigma^2_{a}(x)$ due to variation in environmental variables $\eta$.

\begin{figure*}[htpb]
\centering
\subfloat[target affected by low aleatoric variance \label{fig:ex_3_loss:minvar}]{\includegraphics[width=0.75\textwidth]{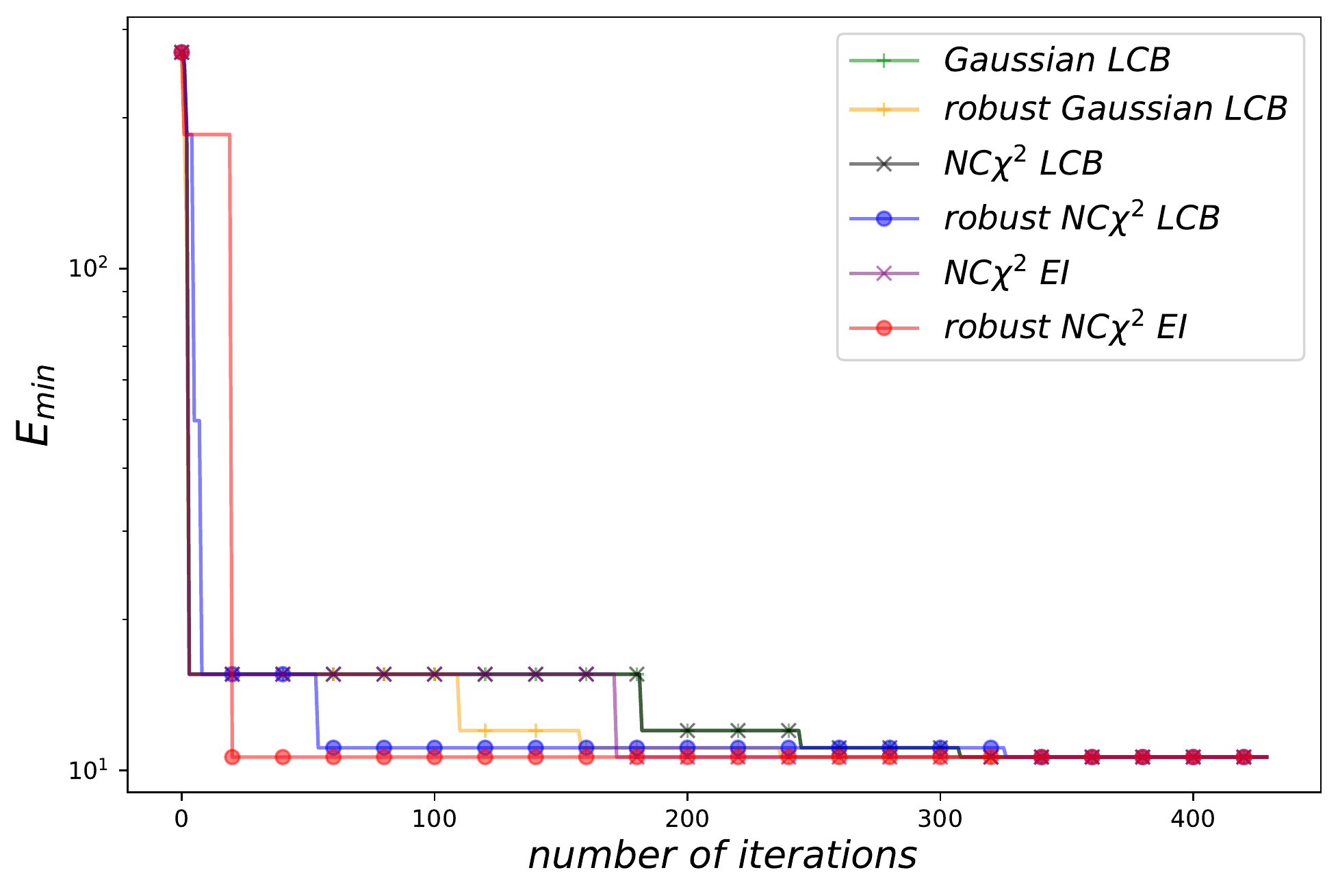}}\hfil
\subfloat[target affected by high aleatoric variance \label{fig:ex_3_loss:maxvar}]{\includegraphics[width=0.75\textwidth]{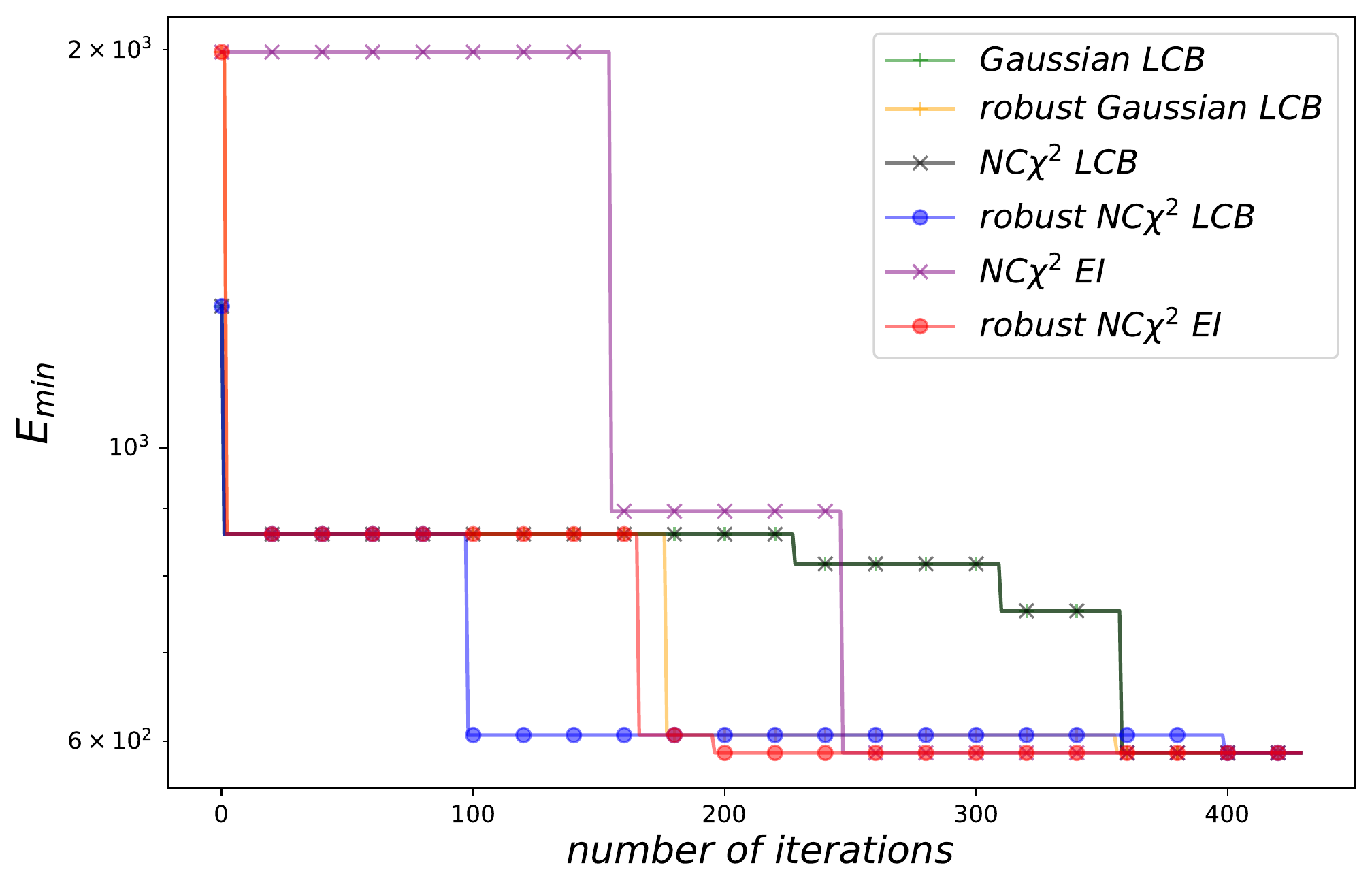}}\hfil 

\caption{Experiment 3: $E_{\min}$ achieved by different methods as a function of the number of sampled data points, i.e., number of optimization steps. 
The $robust~Gaussian~LCB$ method models the variance of the squared error as Gaussian and differentiates between aleatoric and epistemic uncertainty, where a baseline $Gaussian~LCB$ method neglects the differentiation of variances.
The remaining methods estimate variance of squared errors by $NC\chi^2$ distribution and $robust$ methods additionally take advantage of aleatoric variance.
In (\textbf{a}) we choose a target in a region with low aleatoric variance and in (\textbf{b}) we choose a target in a region with high aleatoric variance.
All data points from the pool data set are drawn, therefore, all approaches converge to the overall minimum. }\label{fig:ex_3_loss}
\end{figure*}

We utilize a GP for the mean $m(x)=\expeceta{d_{\max}(x)}$, represented by mean $\mu(x)$ and epistemic variance $\sigma_{e}^2(x)$, and a kernel ridge regression (KRR) model for its empirical variance, representing  $\hat\sigma_a^2(x)$.
GP and KRR are initially trained with two data points, and data points are selected, s.t. the distance in the feature space is maximized, i.e., the first data point holds the lowest possible input values and the second data point the highest possible input values w.r.t. input $x = (\theta_{1}, \theta_{2}, d, h, T)$.
We have found that this selection of initial training data counteracts falling into local optima.
We chose two targets $d_{\max}^\bullet$ from the pool dataset, s.t. one is in a location of low, and one in a location of greater aleatoric variance, cf. Figures \ref{fig:ex_3_loss:minvar} and \ref{fig:ex_3_loss:maxvar}. 
That means that optimal solutions in terms of  $E_{\min}$ are either dominated by the error between target and mean prediction or by the aleatoric variance.

We selected different acquisition functions for our target value BO approach: i) an LCB computed from a Gaussian distribution ($Gaussian~LCB$), where aleatoric and epistemic uncertainties are considered jointly~\cite{hoffer2022gaussian}, ii) an LCB from a Gaussian distribution that maximizes epistemic and minimizes aleatoric uncertainties~\cite{hoffer2022gaussian} and is motivated by~\cite{dai2017stable} ($robust~Gaussian~LCB$), iii) the LCB computed from a non-central $\chi^2$ distribution~\cite{pmlr-v89-uhrenholt19a} ignoring aleatoric uncertainty ($NC\chi^2~LCB$), iv) our LCB~\eqref{eq:LCB} ($robust~NC\chi^2~LCB$), v) the EI computed from a non-central $\chi^2$ distribution~\cite{pmlr-v89-uhrenholt19a} ignoring aleatoric uncertainty ($NC\chi^2~EI$), and vi) our EI~\eqref{eq:EI} ($robust~NC\chi^2~EI$). Note that ii) simultaneously optimizes the expected output w.r.t.\ the target and minimizes the output variance due to aleatoric effects, while iii) and v) explicitly model the distribution of the squared error. Only iv) and vi) both consider the output variance due to aleatoric effects and use the non-central $\chi^2$ distribution to model the error.
We use the data from \cite{hoffer2022gaussian}.

We present optimization results for each iteration in Figure \ref{fig:ex_3_loss:minvar} and \ref{fig:ex_3_loss:maxvar}. 
The number of iterations is equal to the number of pool data, s.t. all approaches converge to the optimal solution.
In Figure \ref{fig:ex_3_loss:minvar}, we chose a target in a location of lower aleatoric uncertainty in the feature space, s.t. the overall best solution $E_{\min}$ is also low, compared to Figure \ref{fig:ex_3_loss:maxvar}, where we chose the target near high aleatoric uncertainty regions, s.t. $E_{\min}$ is greater.
Comparison of Figure \ref{fig:ex_3_loss:minvar} and \ref{fig:ex_3_loss:maxvar} shows that our approach exhibits superior performance w.r.t. convergence in both scenarios. However, in a setting where aleatoric uncertainty is more dominant, see Figure \ref{fig:ex_3_loss:maxvar}, the benefit of distinguishing aleatoric and epistemic uncertainties is more substantial.

Furthermore, evaluation is based on a pool dataset, s.t. in each optimization iteration a data point is drawn from the pool dataset and added to training dataset without laying back.
For each iteration $E(x_i)$ is calculated for the actual training set and minimal values, i.e. $E_{min}$, are used for plotting, see Figure \ref{fig:ex_3_loss}.

Further, by comparing $E_{min}$ and squared error values of optimization results for targets affected by high aleatoric variance (Figure \ref{fig:ex_3_loss:maxvar}), one can discover that approaches, which consider aleatoric effects prefer solutions that minimize $E_{min}$, independent if the squared error is increased. 
For example, our $robust~NC\chi^2~EI$ procedure shows a squared error of about 83.7 at the beginning, which is the same as that of the standard $NC\chi^2~EI$ procedure.
However, after the third optimization iteration, the $robust~NC\chi^2~EI$ prefers a data point with a higher squared error (about 526.7) but a lower aleatoric variance to minimize $E_{min}$ overall.
Neglecting aleatoric effects, the standard $NC\chi^2~EI$ keeps its optimization result at 83.7 until iteration 155, where a better optimization result is found.

We observed that methods that do not use aleatoric uncertainty in the acquisition function spend long time near the selected target $y^{\bullet}$, i.e., $d_{max}^{\bullet}$ as the seemingly 'best' optimization result until the actual optimum is found randomly by further drawing data points.


\section{Discussion and Limitations}\label{sec:discussion}

We proposed acquisition functions for robust BO with the aim that the output of a black box function is close to a target value in the sense of an expected squared error and under the assumption that aleatoric uncertainty due to environmental effects is known or can be learned. We show in our experiments that this assumption is at least approximately compatible with a large set of scenarios, including standard GPs with noisy measurements, GPs with noisy inputs (including cascades of GPs such as deep \cite{Damianou2013} or stacked GPs~\cite{abdelfatah2016environmental,Neumann2009}, in which aleatoric uncertainties can be propagated), and machine learning approaches in which aleatoric and epistemic uncertainties are learned separately. 

While our results show that our acquisition functions outperform classical approaches in the considered task,
our approach and Assumptions~\ref{assumption:noise-free} and~\ref{assumption:environmental} imply certain limitations. In the case of GPs, for example, Assumption~\ref{assumption:noise-free} requires that training and re-training (after obtaining a new data point) relies on noise-free data. When measurement noise is included in the dataset, then the variance of the predictive posterior would include mixed, inseparable components from both epistemic and aleatoric uncertainties, making the separation necessary for our derivations impossible. Similarly, if epistemic and aleatoric uncertainties are represented by other machine learning models, these models must be trained on data that facilitates such a separation. For example, if the mean and the aleatoric uncertainty of the mapping are modeled by a GP and a nonlinear regression model, respectively (as in Section ~\ref{sec:example3}), we need noise-free and unbiased estimates of the mean for training the GP, and noise-free and unbiased estimates of the aleatoric variance for training the regression model. This, in turn, requires taking sufficiently many measurements at each position $x$, such that the mean and the variance of the resulting measurement can be estimated with little error. The advantage of faster convergence of the optimization problem thus has to be traded against the requirement to take multiple (simulation or experimental) measurements. A possible remedy for this limitation could be to allow finite measurement noise up to a magnitude that is small compared to the aleatoric uncertainty, 
or to include a separate mixing term representing inferential uncertainty in the predictive posterior.
Finally, \cite{Ankerman2010} showed that estimates for the aleatoric variance from even very small sample sizes allow for good approximations in the predictive model.

In this work, we measured utility by the expected squared error between the output of the mapping and a given target value, where the expectation is taken over aleatoric effects. Modeling aleatoric and epistemic uncertainties with Gaussian distributions, this operational goal allowed us to derive acquisition functions in closed form. Future research shall extend our work to different practically relevant operational goals. For example, replacing the expected squared error by the probability for an excess error leads to the aim of finding an $x$ that maximizes $\Prob{|y-y^\bullet|<\varepsilon}$, where $\varepsilon$ defines the tolerance level and where the probability is evaluated w.r.t.\ the aleatoric uncertainty. Such a setting may be useful in applications where certain tolerance bands must not be violated.

Until now, we focused on optimization of individual stochastic mappings. However, if one aims to optimize entire manufacturing chains, GP surrogate models can be stacked \cite{Neumann2009}.
In stacked GPs, the output of a previous GP is (a part of) the input for the following GP, and uncertainties are propagated through the entire chain \cite{abdelfatah2016environmental}. Future work shall investigate how aleatoric and epistemic uncertainties can be propagated separately through the stacked GPs, such that our proposed acquisition functions can be utilized for the optimization of entire process chains.

Finally, Bayesian Optimization has also been investigated with respect to scalability. Scalable BO algorithms have been derived for large data sets \cite{Snoek2015,Eriksson2019}, large input dimensions \cite{wang2016bayesian,Daulton2021},  many objectives \cite{Asencio2021} and large output dimensions or many tasks \cite{Hakhamaneshi2021}. A possible future direction is under which circumstances our proposed robustness towards stochastic environmental variables can be extended to these scaling variants. Note however that we made no strong assumptions on the number of environmental variables.

\section{Conclusion}

In this work, we derived a set of acquisition functions for Bayesian target value optimization that is robust against stochastic environmental variables, based on a common Gaussian process surrogate. In contrast to the usual Gaussian distributions of simple minimization/maximization, this leads to non-central chi-square probability density functions for the sought-for optimization objective. This optimization problem was then considered in the presence of aleatoric effects in environmental (non-controllable) variables. We find that knowledge of this aleatoric uncertainty can be leveraged advantageously towards optima that are robust against such stochastic environmental variables. For this, we demonstrate experimentally that estimates or learned models of the aleatoric variance can be sufficient, and that the approach is of particular advantage if aleatoric variance is indeed large. 

Based on the good performance in an alloy billet forging problem, it is speculated that the approach might be useful for broader applications in manufacturing and industrial engineering. Aleatoric uncertainy is, after all, present in many data sets and hence a large class of machine learning or optimization problems.

\subsubsection*{Acknowledgements}
J. G. Hoffer and B. C. Geiger were supported by the project BrAIN - Brownfield Artificial Intelligence Network for Forging
of High Quality Aerospace Components (FFG Grant No. 881039). The project is funded in the framework of
the program 'TAKE OFF', which is a research and technology program of the Austrian Federal
Ministry of Transport, Innovation and Technology. S. Ranftl was supported by University of Technology's LEAD Project 'Mechanics, Modeling and Simulation of Aortic Dissection'. The Know-Center is funded within the Austrian
COMET Program -- Competence Centers for Excellent Technologies -- under the auspices of the
Austrian Federal Ministry of Transport, Innovation and Technology, the Austrian Federal Ministry
of Economy, Family and Youth and by the State of Styria. COMET is managed by the Austrian
Research Promotion Agency FFG.

\bibliography{references}

\clearpage
\appendix
\section{Appendix A: Practical Aspects of the Acquisition Functions}
\label{app:practical}
Maximizing~\eqref{eq:poi} over $x$ is complicated, since the CDF of a non-central $\chi^2$-distribution is not available in closed form. However, it was shown in~\cite{sankaran1959non} that using a non-linear transform, the distribution of $e\sim  NC\chi^2(K,\lambda)$ can be transformed into an approximately Gaussian distribution. Specifically, by setting 
\begin{equation}
    z = \left( \frac{e}{K+\lambda}    \right)^\ell
\end{equation}
with $\ell=1-r_1r_3/3r_2^2$, $r_s = 2^{s-1} (s-1)!(K+s\lambda)$, we get that $z$ has approximately the distribution $\Norm{\alpha,\rho^2}$ with
\begin{subequations}
\begin{align}
    \alpha &= 1+\ell(\ell-1) \left(  \frac{r_2}{2r_1^2} - (2-\ell)(1-3\ell) \frac{r_2^2}{8r_1^4}  \right)\\
    \rho &= \frac{\ell r_2^2}{r_1}  \left(  1- \frac{(1-\ell)(1-3\ell)}{4r_1^2} r_2  \right).
\end{align}
\end{subequations}

By this approximation, we obtain a closed-form approximation of the CDF of, e.g., $e(x)$ as
\begin{equation}
    F_{1,\lambda(x)}(e(x)) \approx \Phi\left(\frac{e(x)-\alpha}{\rho} \right).
\end{equation}

With this, we circumvent the straightforward approach of modelling directly $p_{\mathcal{N}}(e|x,D) = \mathcal{N}(e|\mu(x),\sigma^2(x))$, where $p(e|x)$ is imprecise, because it would be symmetric and with negative support. 

\section{Additional Figures for Experiment 1}
\label{app:example1}

\begin{figure*}

\centering
\subfloat[$\sigma_a= 0.01$ \label{fig:ex_1_LCB:a}]{\includegraphics[height=3.5 cm]{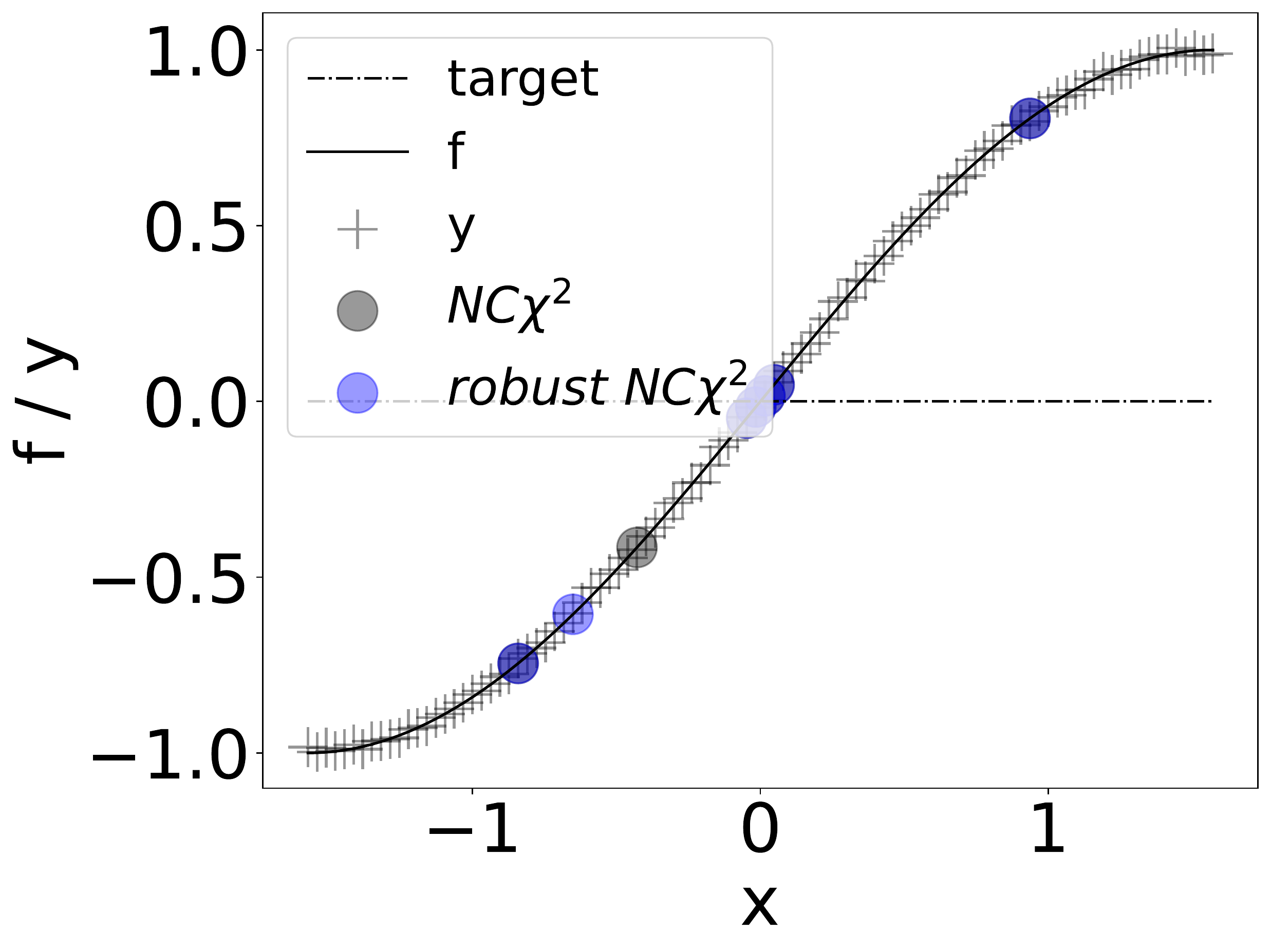}}\hfil
\subfloat[$\sigma_a= 0.1$ \label{fig:ex_1_LCB:b}]{\includegraphics[height=3.5 cm]{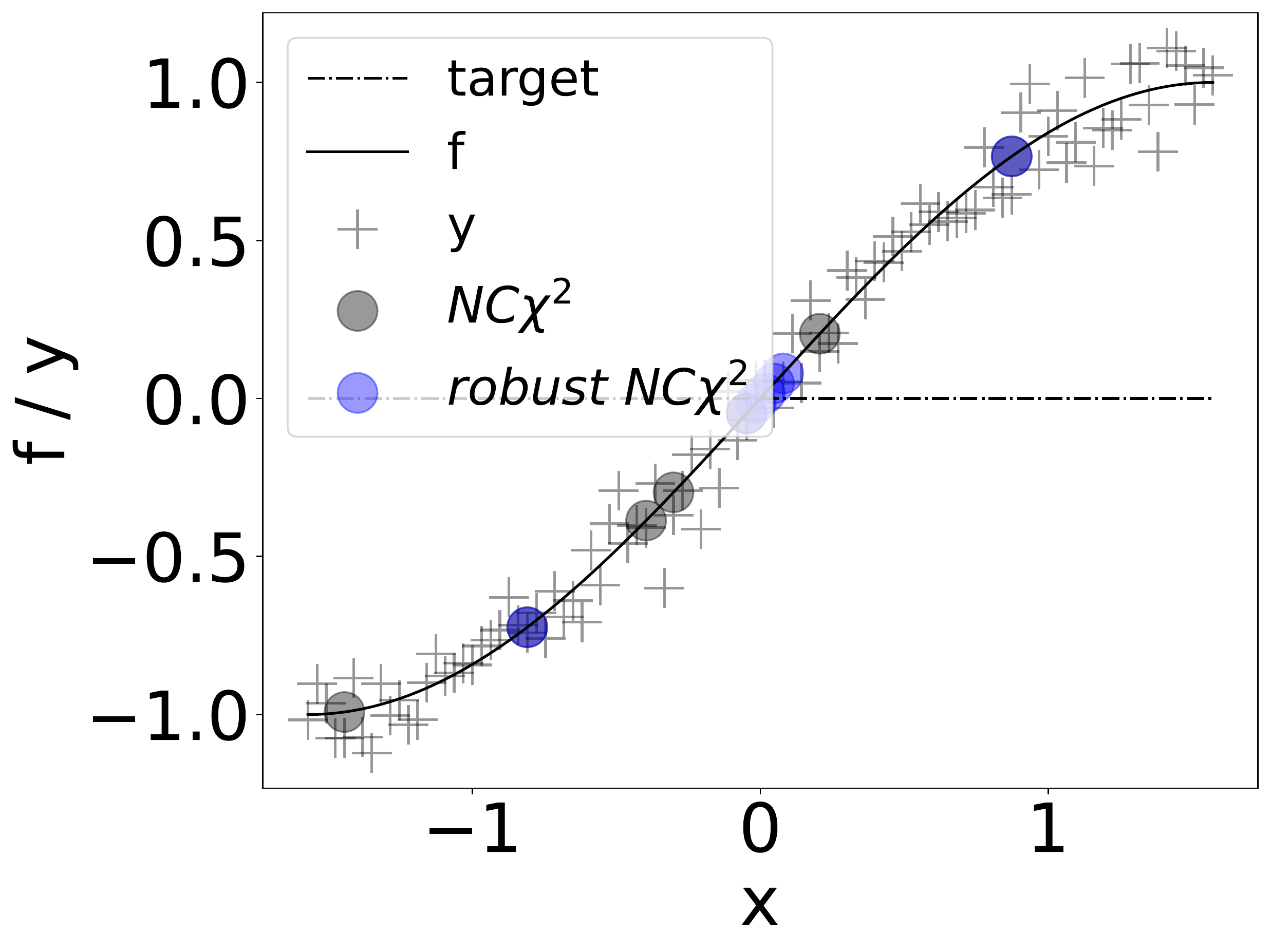}}\hfil 
\subfloat[$\sigma_a= 0.5$ \label{fig:ex_1_LCB:c}]{\includegraphics[height=3.5 cm]{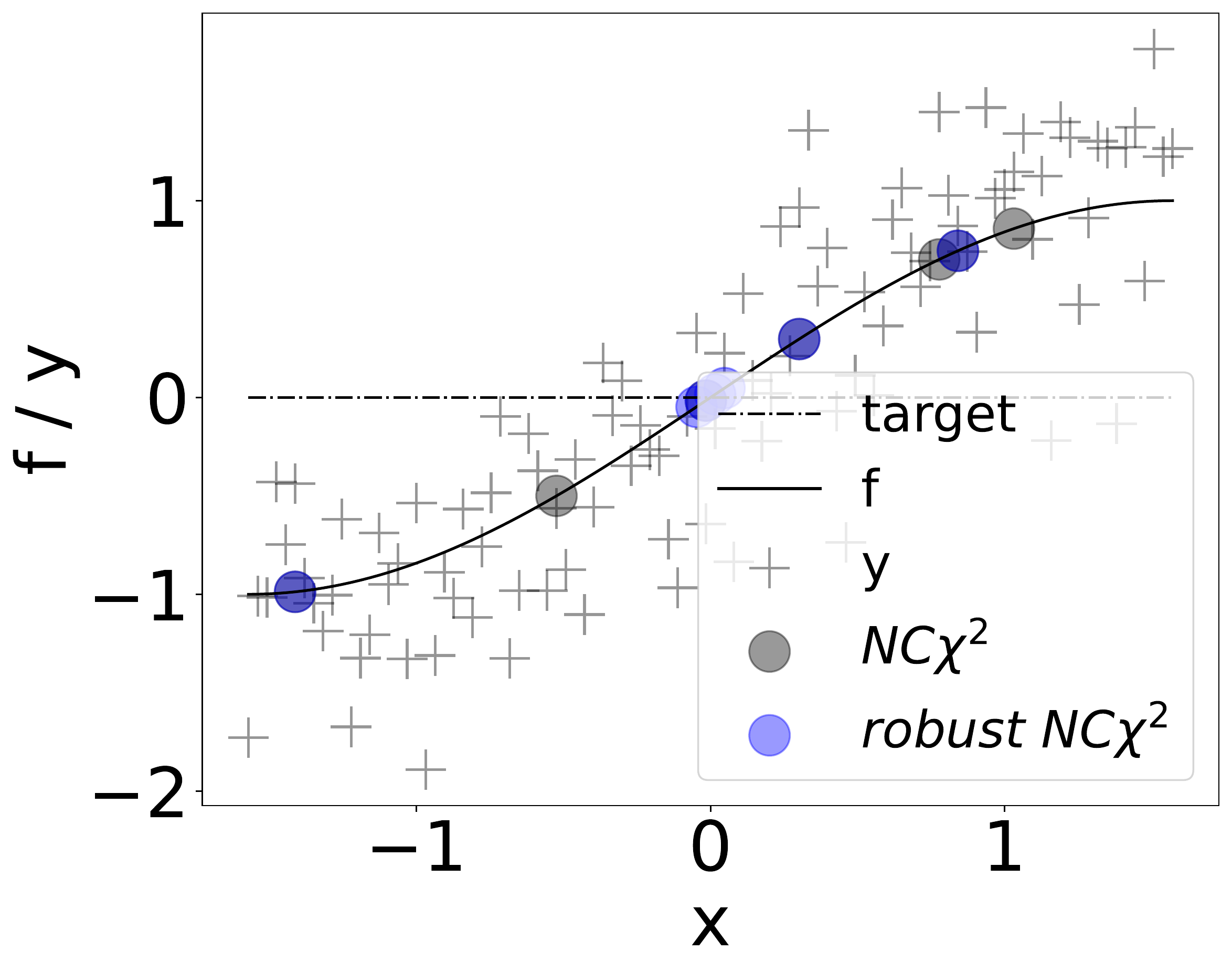}} 

\subfloat[$\sigma_a= 0.01$ \label{fig:ex_1_LCB:d}]{\includegraphics[height=3.5 cm]{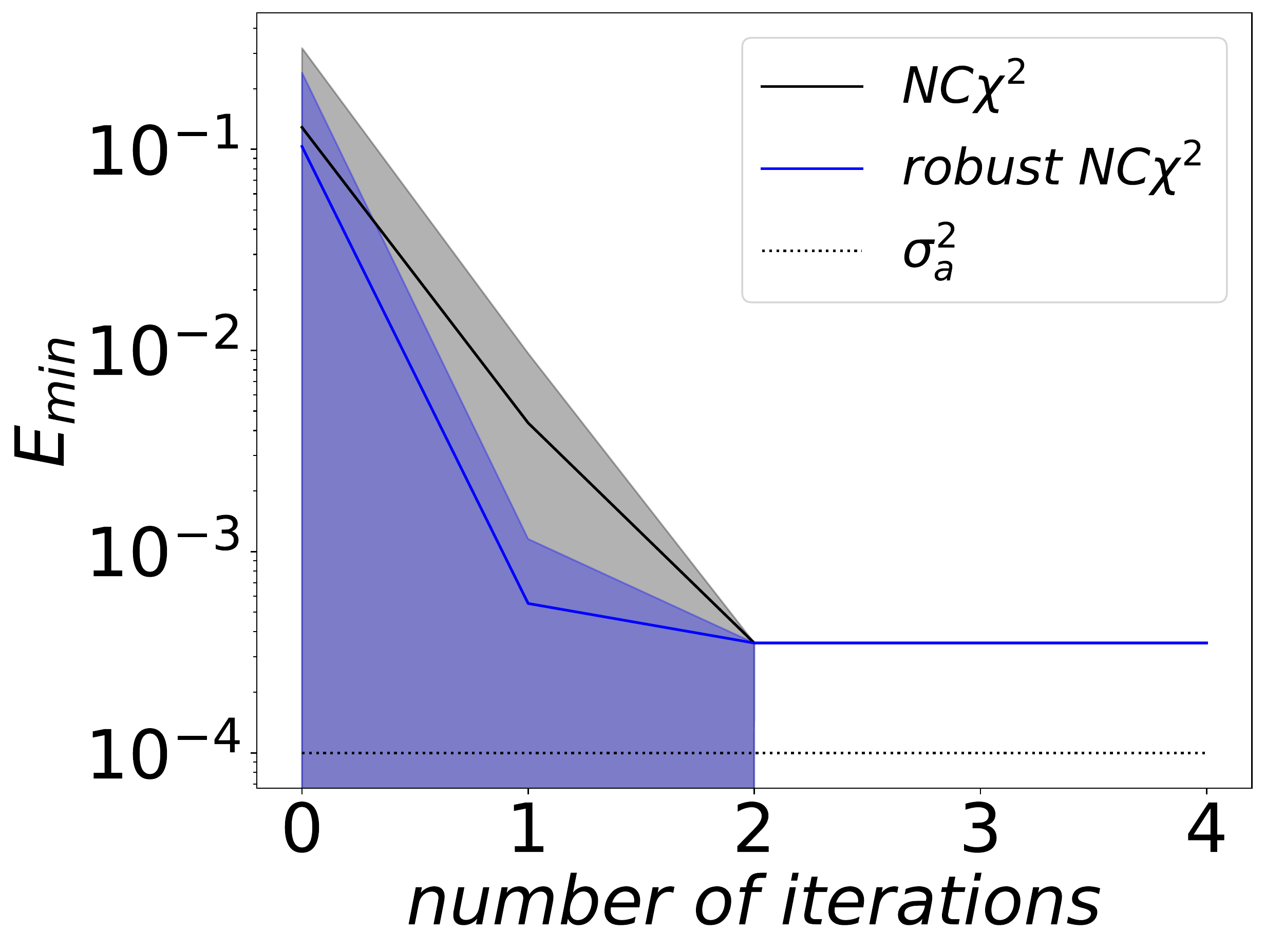}}\hfil   
\subfloat[$\sigma_a= 0.1$ \label{fig:ex_1_LCB:e}]{\includegraphics[height=3.5 cm]{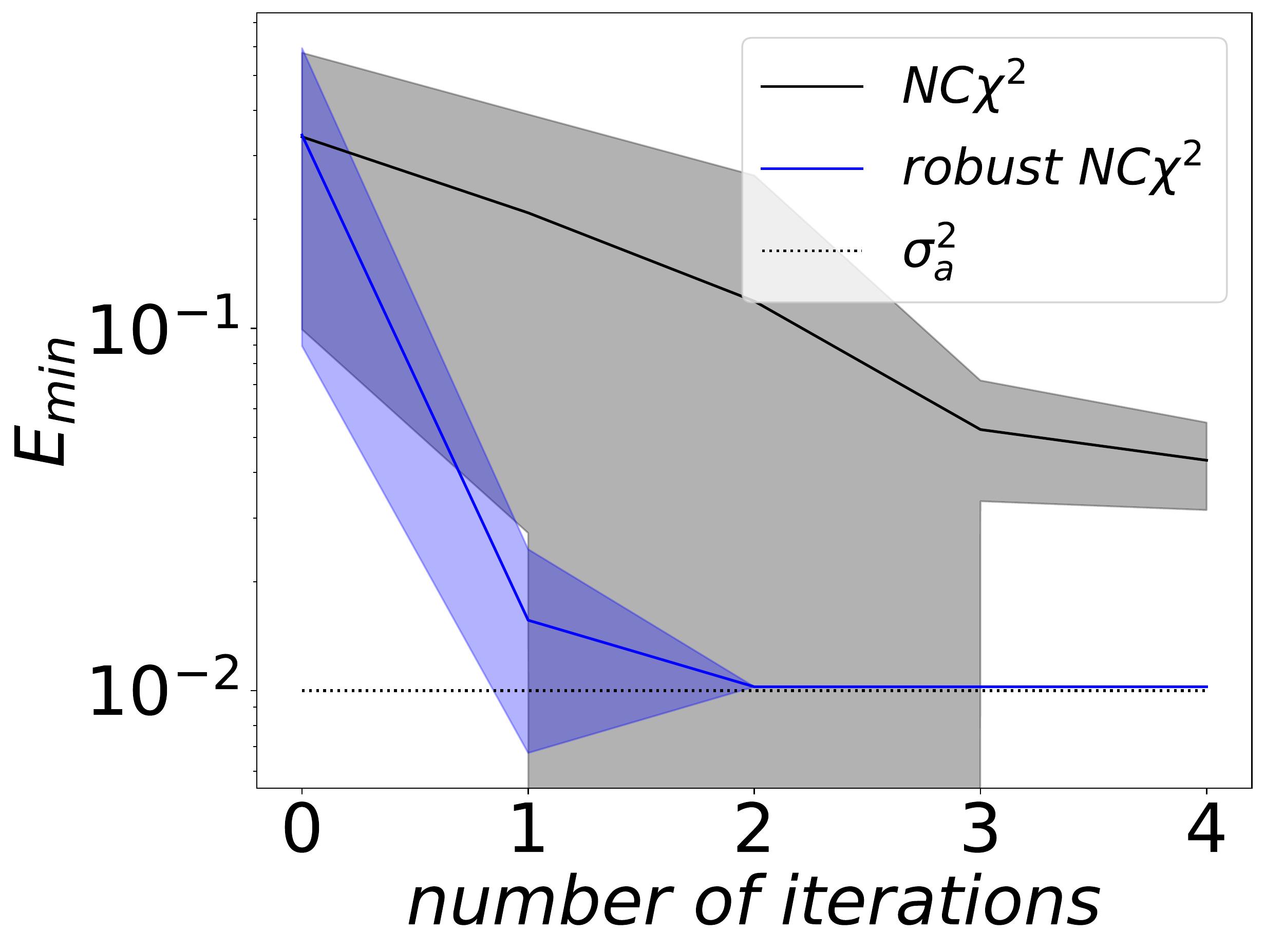}}\hfil
\subfloat[$\sigma_a= 0.5$ \label{fig:ex_1_LCB:f}]{\includegraphics[height=3.5 cm]{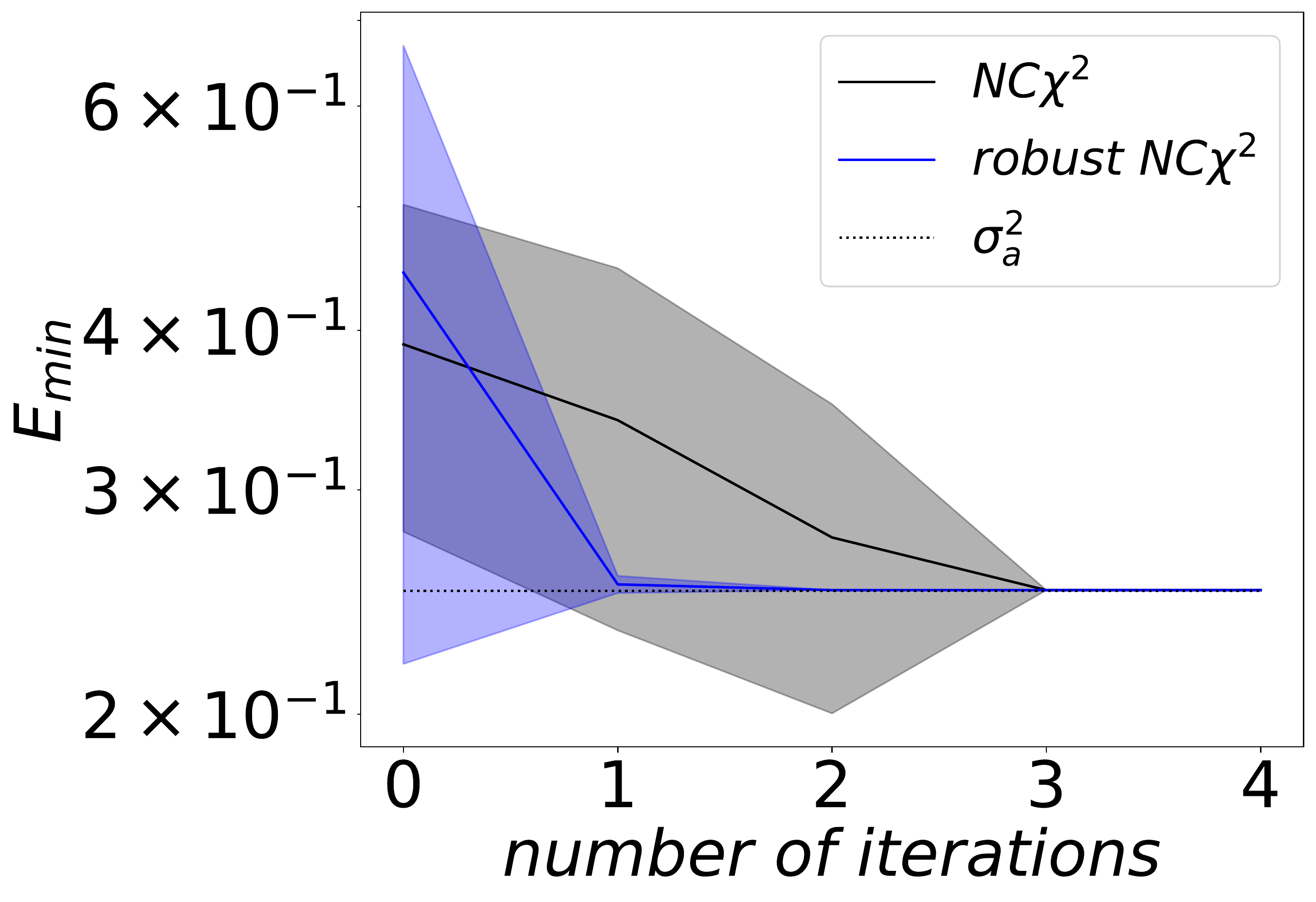}}
\caption{Experiment 1 with LCB~\eqref{eq:LCB}. (\textbf{a})-(\textbf{c}) Scatterplots of the used data with different levels of aleatoric variance $\sigma_a$. Solid lines represent the noise-free function $f$, crosses the observations $y$, dots the sampled candidate positions, and dash-dotted lines the target.  (\textbf{d})-(\textbf{f}) $E_{\min}$ over the number of sampled candidate positions, i.e., number of optimization iterations, for different levels of aleatoric variance $\sigma_a$.}\label{fig:ex_1_LCB}
\end{figure*}

For experiment 1, we did in addition an evaluation by using LCB acquisition function, see Figure \ref{fig:ex_1_LCB}.
Similar to evaluation with EI acquisition function, we can show that by increasing aleatoric uncertainty $\sigma_a$ (Figure \ref{fig:ex_1_LCB:b} - \ref{fig:ex_1_LCB:c}) our $robust ~ NC\chi^2$ outperforms, see Figure \ref{fig:ex_1_LCB:e} - \ref{fig:ex_1_LCB:f}.
In a setting where aleatoric uncertainty is low (Figure \ref{fig:ex_1_LCB:a}) our $robust~NC\chi^2$ acquisition function performs similar to the approach of \cite{pmlr-v89-uhrenholt19a} as expected, as the influence of $\sigma_a$ is rather small in $E_{min}$.

\end{document}